\RequirePackage{fix-cm}
\documentclass{svjour3}                     
\smartqed  
\usepackage[numbers,sort&compress]{natbib}
\usepackage{graphicx}
\usepackage{mathptmx}      
\PassOptionsToPackage{hyphens}{url}\usepackage{hyperref}

\usepackage{url}            
\usepackage{booktabs}       
\usepackage{amsfonts,amsmath}       
\usepackage{nicefrac}       
\usepackage{microtype}      
\DeclareMathOperator*{\argmax}{arg\,max}
\DeclareMathOperator*{\argmin}{arg\,min}
\usepackage{graphicx}
\usepackage{subcaption}
\usepackage{mathtools}
\usepackage{tabularx}
\usepackage{float}
\usepackage{stfloats}
\usepackage{cancel}

\usepackage{wrapfig}
\usepackage{bm}
\usepackage{algorithm}
\usepackage{algorithmic}
\usepackage[title]{appendix}
\usepackage{enumitem}
\usepackage[dvipsnames]{xcolor}

\usepackage{notoccite}

\begin{document}

\title{An Empirical Study of Derivative-Free-Optimization Algorithms for Targeted Black-Box Attacks in Deep Neural Networks  \thanks{This publication is based on work supported by the EPSRC Centre for Doctoral Training in Industrially Focused Mathematical Modelling (EP/L015803/1) in collaboration with New Rock Capital Management.}
}


\author{Giuseppe Ughi \and Vinayak Abrol \and Jared Tanner }


\institute{G. Ughi \at 
            Mathematical Institute, University of Oxford\\
            Andrew Wiles Building,  Radcliffe Observatory Quarter\\
            Oxford, OX2 6GG, United Kingdom \\
              Tel.: +44 1865 273525\\
              \email{ughi@maths.ox.ac.uk} \and
           V. Abrol \at
            Mathematical Institute, University of Oxford\\
            Andrew Wiles Building,  Radcliffe Observatory Quarter\\
            Oxford, OX2 6GG, United Kingdom \\
              Tel.: +44 1865 273525\\
              \email{abrol@maths.ox.ac.uk}           
           \and
           J. Tanner \at
              Mathematical Institute, University of Oxford\\
            Andrew Wiles Building,  Radcliffe Observatory Quarter\\
            Oxford, OX2 6GG, United Kingdom \\
              Tel.: +44 1865 273525\\
              \email{tanner@maths.ox.ac.uk}           
}

\date{Received: date / Accepted: date}

\maketitle

\begin{abstract}
    We perform a comprehensive study on the performance of derivative free optimization (DFO) algorithms for the generation of targeted black-box adversarial attacks on Deep Neural Network (DNN) classifiers assuming the perturbation energy is bounded by an $\ell_\infty$ constraint and the number of queries to the network is limited.  This paper considers four pre-existing state-of-the-art DFO-based algorithms along with the introduction of a new algorithm built on BOBYQA, a model-based DFO method. We compare these algorithms in a variety of settings according to the fraction of images that they successfully misclassify given a maximum number of queries to the DNN. 
    The experiments disclose how the likelihood of finding an adversarial example depends on both the algorithm used and the setting of the attack; algorithms limiting the search of adversarial example to the vertices of the $\ell^\infty$ constraint work particularly well without structural defenses, while the presented BOBYQA based algorithm works better for especially small perturbation energies. This variance in performance highlights the importance of new algorithms being compared to the state-of-the-art in a variety of settings, and the effectiveness of adversarial defenses being tested using as wide a range of algorithms as possible. 
    
\end{abstract}

\section{Introduction}

Deep Neural Networks (DNNs) achieve state-of-the-art performance on a growing number of applications such as acoustic modelling~\cite{hinton2012deep}, image classification~\cite{he}, and fake news detection~\cite{monti2019fake} to name but a few. Alongside their growing application, there is a literature on the robustness of deep networks which shows that it is often possible to subtly perturb the input image of a DNN in order to degrade its performance; these perturbations are referred to as adversarial examples \cite{goodfellow2014,szegedy2013}. For example, see \cite{dalvi,eykholt2017robust,kurakin,sitawarin,yuan} where road signals are perturbed so as to be wrongly interpreted by self driving cars that analyze images of them with DNNs.
Methods to generate these adversarial examples are classified according to two main criteria \cite{yuan}:
\begin{description}
    \item[\textbf{Adversarial Specificity}] establishes what the aim of the adversary is. In \textit{non-targeted} attacks, the method perturbs the image in such a way that it is misclassified into any category other than the original one. While in \textit{targeted} settings, the adversary specifies a category into which an image should be misclassified. 
  
     \item[\textbf{Adversary's Knowledge}] defines the amount of information available to the adversary. In \textit{White-box} settings the adversary has complete knowledge of  the network architecture and weights, while in the \textit{Black-box} setting the adversary is only able to obtain the pre-classification output vector. The White-box setting allows for the use of gradients of a missclassification objective to efficiently compute the adversarial example \cite{carlini,Chen2018ead,goodfellow2014},  while the  same optimization formulation of the Black-box setting requires use of a derivative free approach \cite{Alzantot,chen,ilyas2018black,narodytska2017}.
     
\end{description}

In this work we consider the targeted black-box setting. In particular we follow \cite{chen} where:
\begin{itemize}
     \item  the \textit{perturbation}, which causes the network to change the classification, is bounded in magnitude by a specified $\ell^{\infty}$-norm, $\varepsilon_\infty$, i.e. each pixel in the image cannot be perturbed by more than $\varepsilon_\infty$;
    \item the \textit{number of queries} to the DNN needed to generate a targeted adversarial example should be as small as possible.
\end{itemize}

\begin{figure}[t]
\centering
    \begin{subfigure}{.85\textwidth}
      \centering
      \includegraphics[width=.99\linewidth]{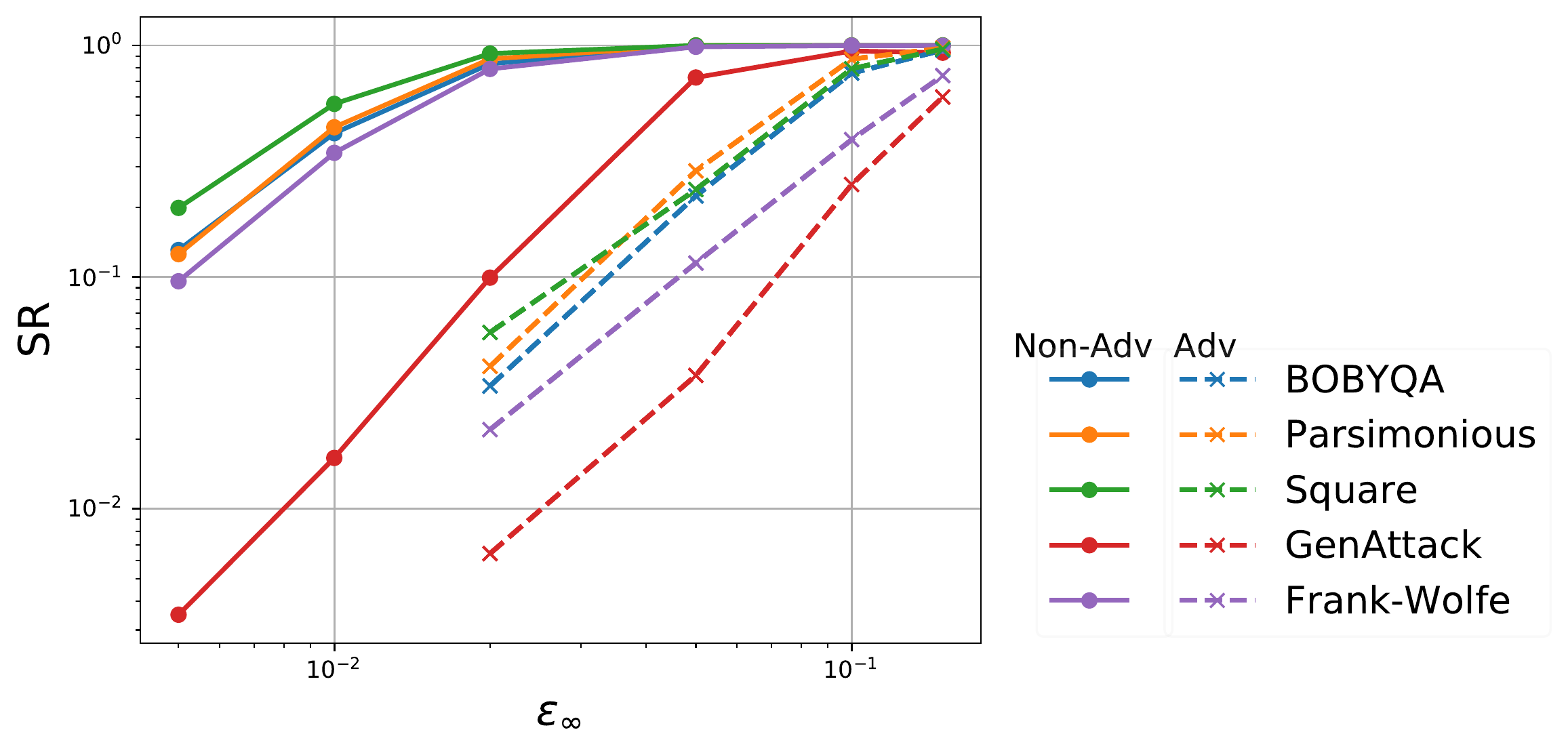}
      \caption{Cifar10}
    \end{subfigure}\\
    \begin{subfigure}{.85\textwidth}
      \centering
      \includegraphics[width=.99\linewidth]{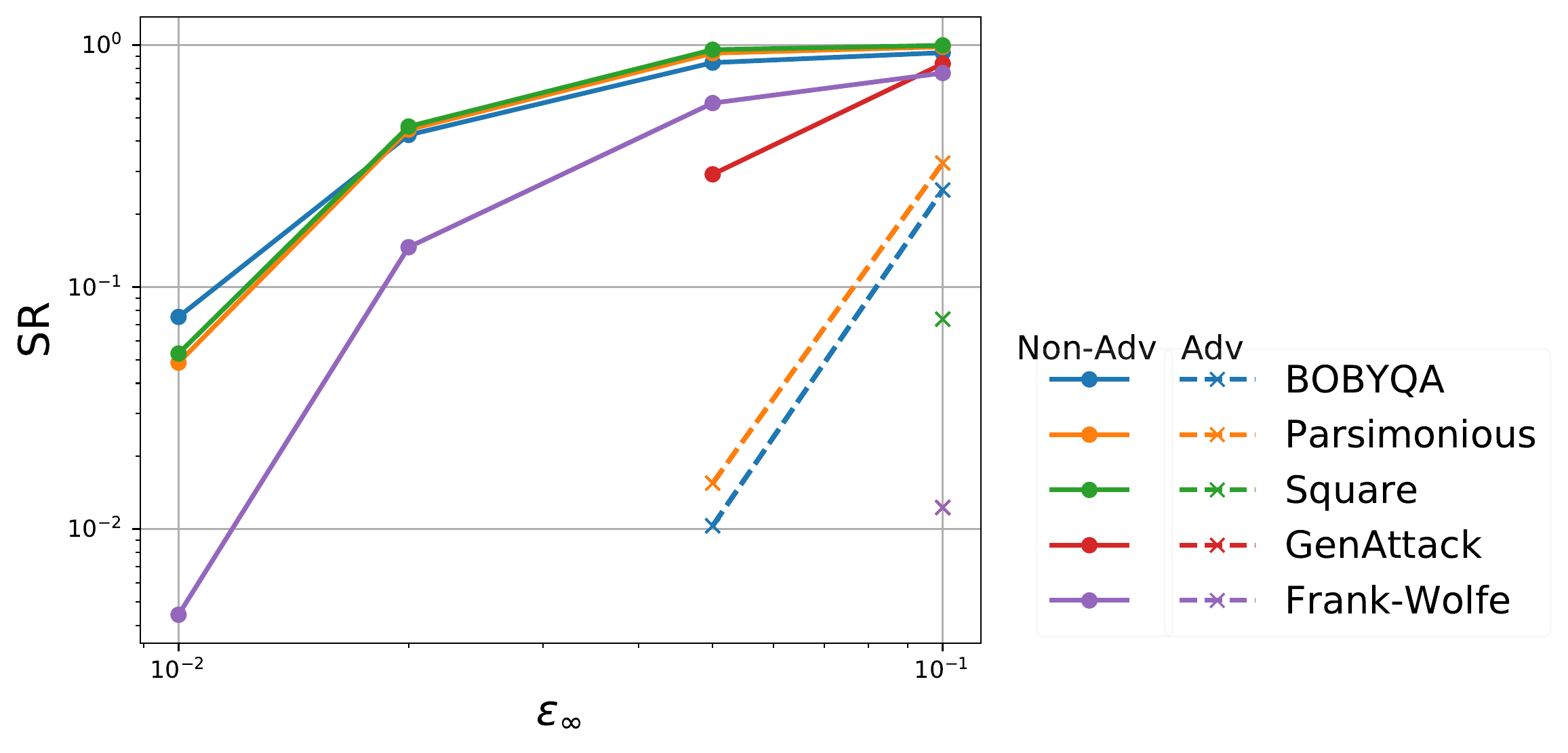}
      \caption{ImageNet}
    \end{subfigure}%
\caption{The success rate (SR) of targeted attacks as a function of the perturbation's allowed $\ell^\infty$ magnitude for algorithms: GenAttack \cite{Alzantot}, Parsimonious \cite{COMBI}, Square \cite{andriushchenko2019square}, Frank-Wolfe \cite{chen2020frank}, and the BOBYQA based algorithm introduced here.  Specifically for a ResNet50 network trained either on the CIFAR10 (a) or the ImageNet (b) dataset with (Adv) and without (Non-Adv) the defense by MadryLab \cite{robustness}. An attack is considered successful if the method found the targeted adversarial example with less than 3'000 or 15'000 queries to the network trained on CIFAR and ImageNet dataset, respectively; 
Results for the case SR=0 i.e., when no perturbations were successful, are excluded from the plot.
}
\label{fig:compa}
\end{figure}

The Zeroth-Order-optimization (ZOO) algorithm proposed in~\cite{chen} describes a Derivative Free optimization (DFO) method for computing adversarial examples in the black-box setting using a coordinate descent optimization method. At the time this was a substantial departure from previous black-box algorithms which trained a proxy DNN and then employ gradient based white-box attacks on the proxy network \cite{papernot,tu2018}. It was demonstrated in~\cite{chen} that these algorithms are especially effective when numerous adversarial examples are computed, but become less efficient when an individual adversarial examples is considered. Following the introduction of ZOO, there have been numerous improvements using other model-free DFO based approaches, see  for example \cite{al2019there,Alzantot,andriushchenko2019square,chen2020frank,ilyas2018black,ilyas2019prior,COMBI}.
Many of these algorithms were developed in parallel, and so have not yet been bench-marked in a consistent setting, e.g. on the same network.  

In this article, we present two frameworks for comparative evaluation of the existing algorithms that claim to have the fewest number of DNN queries to generate a successful attack. These are: GenAttack \cite{Alzantot} which is based  on a genetic direct-search method; Parsimonious algorithm \cite{COMBI}, based on a combinatorial direct-search method on the vertices of the perturbation domain; the Square algorithm \cite{andriushchenko2019square}, based on a randomized direct-search method on the vertices of the perturbation domain; and the Frank-Wolfe algorithm \cite{chen2020frank} based on a momentum mechanism that approximates the gradient via finite differences. We also introduce a new algorithm built on a model-based DFO method~\cite{BOBYQA_Neurips}. In particular, we consider the \textit{Bounded optimization BY Quadratic Approximation} (BOBYQA)~\cite{powellbobyqa} model-based DFO method which explicitly develops pseudo models to approximate the loss function in the optimization problem and then minimizes the loss function using methods from continuous optimization on the generated models. The aforementioned list of algorithms covers the leading classes of DFO algorithms for limited function evaluations, see e.g.,~\cite{conn2009introduction,larson_menickelly_wild_2019} for recent reviews of DFO methods. The two frameworks are structured as follows: 
\begin{enumerate}
    \item In the first setting we consider attacks on DNNs trained on CIFAR10 and ImageNet datasets, with or without the adversarial defense by MadryLab \cite{robustness}; this is the canonical setup for the comparison of black-box attacks that was considered in previous literature. We illustrate in Figure \ref{fig:compa} a measure of how the performance of the considered algorithms compare, while further refined measures of comparison are included in Section \ref{sec:exp}. We observe that the algorithms that limit the optimization domain to the $\ell^\infty$ perturbation boundary, i.e. the Parsimonious and Square algorithms, are consistently the most effective. In particular, the Square algorithm achieves the highest Success Ratio (SR) with a fixed maximum number of queries, except for when the DNNs have been adversarially trained, and the Parsimonious algorithm achieves the highest SR when a network is trained with the MadryLab defense. However, these results are relative to the current state-of-the-art defense in a field which is in continuous development \cite{dhillon2018stochastic,ijcai2019-833} and newly proposed methods usually have a varying effect on the different attacking algorithms; for example the MadryLab defense \cite{robustness} that we consider is most effective on Square algorithm in the ImageNet case.
    \item In the second framework, the algorithms are allowed to perturb only a fraction of the pixels in the input; this is especially inspired by the structural defenses that transform the input in the wavelet space \cite{guo2018countering}. This framework allows us to understand the sensitivity of different algorithms to choices such as initialization, experimental protocol, dataset, and adversarial training. Our results demonstrate that the Parsimonious, Square, and BOBYQA based algorithms alternatively perform the best for different maximum perturbation energies.  
\end{enumerate}  
The results in this paper show that the most likely algorithm to find an adversarial example varies according to the considered setting; the type of dataset, the defense, and the perturbation energy bound have a varying impact on the different algorithms. As a consequence of these experiments, new algorithms should be compared to the state-of-the-art in a variety of settings as done here, and the effectiveness of an adversarial defense should be tested with a variety of algorithms, including the BOBYQA based algorithm introduced in this paper.

The outline of the paper is as follows: 
in Section~\ref{sec:opt} we present how an adversarial example is generated by solving an optimization problem, and how DFO methods fit in this context. We also introduce the model-based BOBYQA algorithm. In Section~\ref{sec:scalability} we present two popular techniques used in existing methods to improve the efficiency and scalability to high dimensional inputs. Section~\ref{sec:exp} presents the experimental setup and a comparative analysis of existing algorithms along with a focus on our proposed BOBYQA based algorithm. We close with some concluding remarks in Section~\ref{sec:summary}.

\section{Adversarial Examples Formulated as an optimization Problem}
\label{sec:opt}

In classification tasks, a DNN outputs a vector whose length is equal to the number of classes and the DNN parameters are trained to match the maximum element of the given output to the correct class of the input.  Adversarial perturbations are obtained by modifying the input in such a way that the maximum element of DNN output corresponds to a target class different from the original one. 

Consider a classification operator $F:\mathcal{X}\rightarrow\mathcal{C}$ from input space $\mathcal{X}$ to output space $\mathcal{C}$ of classes. A targeted adversarial perturbation $\bm{\eta}$ to an input $\textbf{X} \in \mathcal{X}$ has the property that it changes the classification to a specified target class $t$, i.e $F(\textbf{X})= c$ and ${ F(\textbf{X}+\bm{\eta})=t \neq c}$.

Following the formulation in \cite{Alzantot}; given an input space $\mathcal{X}=[l,u]^n$, with $l$ and $u$ being respectively the minimum and maximum values of the interval in which the pixels may vary, an output space ${\mathcal{C}= \{1,\hdots, n_c\}}$, where $n_c$ is the number of classes,  a maximum energy budget $\varepsilon_{\infty}$, and a suitable loss function $\mathcal{L}$, then the task of computing the adversarial perturbation $\bm{\eta}$ can be cast as an optimization problem such as
\begin{align}
    \min_{\bm{\eta}} \;& \mathcal{L}(\textbf{X},\bm{\eta})\label{eq:opt_prob_beg}\\
    \text{s.t.} \;& \|\bm{\eta} \|_\infty \leq \varepsilon_\infty; \nonumber\\
    & [\textbf{X} + \bm{\eta}]_j \geq l \;\quad\quad\quad \forall j \in {1,...,n} \nonumber \\
    & [\textbf{X} + \bm{\eta}]_j \leq u \;\quad\quad\quad \forall j \in {1,...,n} \nonumber \label{eq:opt_prob_end}
\end{align}
where the final two inequality constraints are due to the perturbed image being still an image, i.e. $(\textbf{X}+\bm{\eta})\in \mathcal{X}$.  Denoting the pre-classification output vector by $f(\textbf{X})$, i.e. ${F(\textbf{X}) = \argmax f(\textbf{X})}$, then the misclassification of $\textbf{X}$ to target label $t$ is achieved by $\bm{\eta}$ if $f(\textbf{X} + \bm{\eta})_t \geq \max_{j\neq t} f(\textbf{X} + \bm{\eta})_j$. As demonstrated in~\cite{Alzantot,carlini,chen}, in this study we consider the following loss function for computing $\bm{\eta}$ in \eqref{eq:opt_prob_beg}
\begin{equation}\label{eq:loss}
    \mathcal{L}(\textbf{X},\bm{\eta}) = \log\left(\Sigma_{j\neq t}f(\textbf{X}+\bm{\eta})_j\right) - \log\left(f(\textbf{X}+\bm{\eta})_t\right).
\end{equation}
Not having access to the internal parameters of the DNN, the gradient of the loss over the input space cannot be readily computed and instead the adversarial perturbation is found using specially adapted DFO algorithms.

\subsection{Derivative Free optimization for Adversarial Examples}

Derivative Free optimization is a well developed field with numerous classes of methods, see \cite{conn2009introduction} and \cite{larson_menickelly_wild_2019} for reviews on DFO principles and algorithms.  Example classes of such methods include: direct search methods such as simplex, model-based methods, hybrid methods such as finite differences or implicit filtering, as well as randomized variants of the aforementioned and methods specific to convex or noisy objectives.  For the generation of adversarial examples, the algorithms that we consider rely on three types of DFO methods: 
\begin{itemize}
    \item those where the gradient is computed via finite differences, either by sampling all the canonical directions as in ZOO attack \cite{chen} or random directions as in the Frank-Wolfe algorithm \cite{chen2020frank};
    \item those where the solution is thought to be in one of the vertices of the $\ell^{\infty}$ domain, i.e. $\bm{\eta}_i \in \{-\varepsilon_\infty, \varepsilon_\infty \}$ for any $i$. The Parsimonious algorithm \cite{COMBI} implements a combinatorial direct-search within the different possible vertices, initializing the perturbation to $-\varepsilon_\infty$ for all the pixels and then switching collections of them to $+\varepsilon_\infty$, when such an action decreases the loss function. The Square algorithm \cite{andriushchenko2019square} instead implements a randomized direct-search method where square blocks of pixels are iteratively perturbed to be either $+\varepsilon_\infty$ or $-\varepsilon_\infty$;
    \item those where a direct search over the perturbation domain is performed using a genetic method such as GenAttack \cite{Alzantot}.
\end{itemize}
The optimization formulation in (\ref{eq:opt_prob_beg}) is amenable to virtually all DFO methods, making it unclear which of the methods would be most effective in this context. Further, model-based methods are notably missing from the aforementioned list. Thus for completeness, we introduce an algorithm relying on a model-based method; specifically, BOBYQA is considered given its proven effectiveness in solving complex problems such as climate modelling~\cite{Climate}.

\subsection{Model-Based DFO}
Given a set of $q$ samples $\mathcal{Y} = \{\textbf{y}^1,...,\textbf{y}^q\}$ with $\textbf{y}^i$ $\in \mathbb{R}^n$, model-based DFO methods start by identifying the minimizer of the objective among the samples at iteration $k$, $\textbf{x}^k =\argmin_{\textbf{y}\in \mathcal{Y}} \mathcal{L}(\textbf{y})$. Following this, a model for the objective function $\mathcal{L}$ is constructed, typically centered around the minimizer. In its simplest form one uses a polynomial approximation to the objective, such as a quadratic model centered in $\textbf{x}^k$
\begin{equation}\label{eq:Model}
    m_k(\textbf{x}^k + \textbf{p}) = a_k + \textbf{c}_k^\top \textbf{p} + \frac{1}{2}\textbf{p}^\top \textbf{M}_k \textbf{p},
\end{equation}
with $a_k\in\mathbb{R}$, $\textbf{c}_k$, $\textbf{p}\in\mathbb{R}^n$, and $\textbf{M}_k\in \mathbb{R}^{n\times n}$ being also symmetric. In a white-box setting one would set $\textbf{c}_k = \nabla \mathcal{L}(\textbf{x}^k)$ and $\textbf{M}_k= \nabla^2 \mathcal{L}(\textbf{x}^k)$, but this is not feasible in the black-box setting as we do not have access to the derivatives of the objective function. Thus at each iteration $k$, the parameters $a_k$, $\textbf{c}_k$ and $\textbf{M}_k$ are usually defined by imposing interpolation conditions

\begin{equation}\label{eq:interpolation}
 m_k(\textbf{y}^i) = \mathcal{L}(\textbf{y}^i) \quad \forall i \in 1,2,\ldots,q,
\end{equation}
and when $q<1 + n + n(n+1)/2$ (i.e. the system of equations is under-determined) other conditions are introduced according to which method is considered. The objective model \eqref{eq:Model} is considered to be a good estimate of the objective in a neighborhood referred to as a trust region.  Once the model $m_k$ is generated, the update step $\textbf{p}$ is computed by solving the trust region problem
\begin{align}\label{eq:Hessain}
    \min_\textbf{p} \quad & m_k(\textbf{x}_k + \textbf{p})\\ \text{s.t.} & \quad\| \textbf{p}\| \leq \Delta,\notag
\end{align} 
where $\Delta$ is the radius of the region where we believe the model to be accurate, for more details see \cite{nocedal}. The new point $\textbf{x}_k + \textbf{p}$ is added to $\mathcal{Y}$ and a prior point is potentially removed. In this paper, we consider an exemplary model-based method called BOBYQA.

\subsubsection{BOBYQA}
The \textit{Bound Optimization BY Quadratic Approximation} (BOBYQA) method, introduced in \cite{powellbobyqa}, updates the parameters of the model $a,\textbf{c},$ and $\textbf{M}$, in each iteration in such a way as to minimize the change in the quadratic term $\textbf{M}_k$ between iterates while otherwise fitting the sample values:
\begin{align}
    \min_{a_k,\textbf{c}_k,\textbf{M}_k} & \| \textbf{M}_k - \textbf{M}_{k-1}\|_F^2 \; \label{eq:update_model}\\ 
    \text{s.t.} \quad &\;   m_k(\textbf{y}^i) = \mathcal{L}(\textbf{y}^i), \;\quad \quad \forall i \in 1,2,\ldots,q, \notag
\end{align}
with $n+1 < q <  1 + n + n(n+1)/2$ and $\textbf{M}_k$ initialized as the zero matrix. When the number of parameters $q = n+1$ then the model is considered as linear with $\textbf{M}_k$ set as zero. Every time a new query is done, the sample which is the least important geometrically is removed from $\mathcal{Y}$, thus keeping the dimension of $\mathcal{Y}$ fixed.

\section{Improving Efficiency and Computational Scalability}
\label{sec:scalability}
Because of the high number of pixels in the input images, the generation of adversarial examples involves solving a high dimensional problem, which makes the use of any DFO method impractical;  for instance, the application of the BOBYQA method requires the solution of \eqref{eq:update_model} which scales in memory allocation at least quadratically with the input dimension, and thus is computationally too expensive. Consequently, the implementation of DFO based adversarial algorithms relies on strategies to reduce the dimensionality of the problem, this improves the computational scalability along with the efficiency, as demonstrated experimentally. Instead of solving \eqref{eq:opt_prob_beg} for $\bm{\eta}\in\mathbb{R}^n$ directly,  the DFO based algorithms consider variations of the domain sub-sampling and/or hierarchical liftings techniques. Domain sub-sampling iteratively sweeps over batches of $b\ll n$ variables, while hierarchical lifting clusters and perturbs variables simultaneously, as described in following sections.

\subsection{Domain Sub-Sampling}

\begin{figure}[t!]
    \centering
    \includegraphics[width=.75\linewidth]{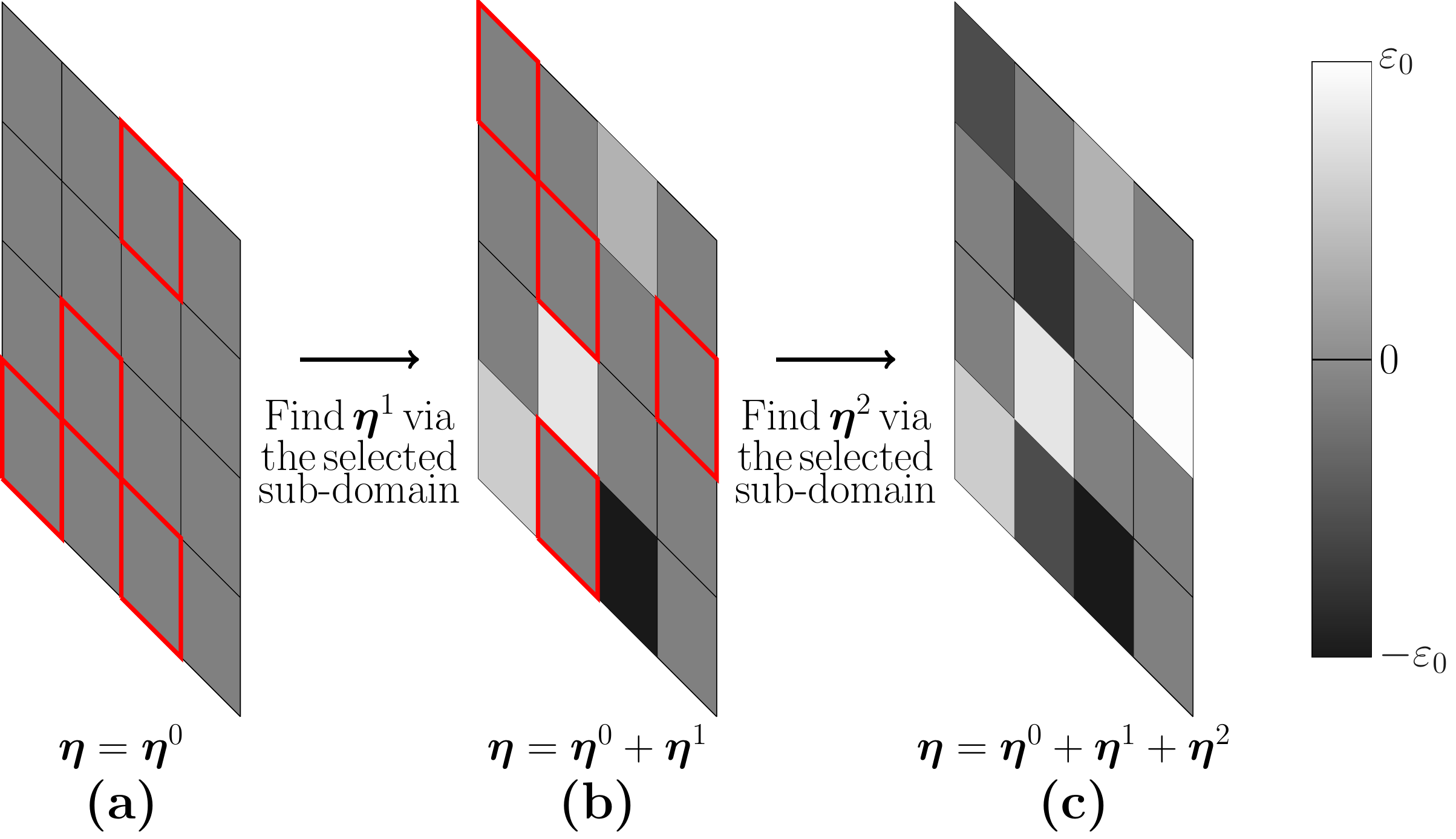}
    \caption{Example of how the perturbation $\bm{\eta}$ evolves through the iterations when an image in $\mathbb{R}^{4\times4}$ is attacked. In (a) the perturbation is $\bm{\eta} = \bm{\eta}^0$ and a sub-domain of $b=4$ pixels (in red) is selected. Once the optimal perturbation $\bm{\eta}^1$ in the selected sub-domain is found, the perturbation is updated in (b) and a new sub-domain of dimension $b$ is selected. The same is repeated in (c).}
    \label{fig:Sub-Domain}
\end{figure}

The simplest version of domain sub-sampling consists of partitioning the input dimension into smaller disjoint domains and optimizing the loss function in each of them sequentially. This is,  in an $n$ dimensional problem, one considers $k= \lceil n/b \rceil$ sets of integers,  $\{\Omega^j\}_{j=1}^k$, of  size $b\ll n$ which are disjoint and which cover all of $[n]$. Then \eqref{eq:opt_prob_beg} is solved sequentially on the dimensions identified by the sets $\Omega^j$. This is possible since the optimization domain is box like, i.e. $\bm{\eta}\in[l,u]^n$, and each dimension's bound is independent from the others. Formally, rather than solving \eqref{eq:opt_prob_beg} for $\bm{\eta}\in\mathbb{R}^n$ directly, for each of  $j=1,\ldots,k$ one {\em sequentially} solves for the $\bm{\eta}^j\in\mathbb{R}^n$ variables which are only non-zero for entries in $\Omega^j$.  The resulting sub-domain perturbations $\bm{\eta}^j$ are then summed to generate the full  perturbation     
$\bm{\eta} = \sum_{j=1}^k \bm{\eta}^j$, see Figure \ref{fig:Sub-Domain} as an example.  That is, the optimization problem (\ref{eq:opt_prob_beg}) is adapted to repeatedly looping over $j=1,\ldots,k$: 
\begin{align}
    \min_{\bm{\eta^j}} \;\; & \mathcal{L}\left(\textbf{X}+\sum_{h\neq j}\bm{\eta}^{\ell},\bm{\eta}^j\right) 
    \; \label{eq:opt_sub_samp}\\
    \;\text{s.t.} \quad &\left\|\sum_{h=1}^k \bm{\eta}^{h} \right\|_\infty \leq \varepsilon_\infty;  \nonumber \\
    & \left[\textbf{X} +\sum_{h=1}^k\bm{\eta}^{h}  \right]_r \geq l \;\quad\quad\quad \forall r \in \Omega^j; \nonumber \\
    &\left[\textbf{X} + \sum_{h=1}^k\bm{\eta}^{h} \right]_r \leq u \;\quad\quad\quad \forall r \in \Omega^j,\nonumber 
    \label{eq:opt_prob_proj_end}
\end{align}
where the sets $\{\Omega^j\}_{j=1}^k$ are usually computed again once $j$ is equal to $k$, and the sub-domain perturbations $\bm{\eta}^j$ are initialized as null.

We identified three possible ways of selecting the sub-domains $\{\Omega^j\}_{j=1}^k$;
\begin{itemize}
    \item In \textit{Random Sampling} one considers at each iteration a different random sub-samplings of the domain, i.e. $k=1$. The ZOO algorithm used this kind of sampling \cite{chen}.
    \item In \textit{Ordered Sampling} one generates a random disjoint partitioning of the domain, i.e. $k=\lceil n/b \rceil$ and {$\Omega_j \cap \Omega_l = \emptyset$} for any $j$ and $l$. A new partitioning is generated when each variable has been optimized over once. This sampling is implemented in the Parsimonious algorithm.
    \item In \textit{Variance Sampling} one still generates a a random disjoint partitioning of the domain, but chooses the sub-samplings sets $\{\Omega^j\}_{j=1}^{k}$ in order to optimize over the dimensions that have highest local variance in intensity first. Specifically, the variables are ordered by the variance in intensity among the 8 neighboring variables (e.g. pixels) in the same color channel of the input $\textbf{X}$. The sets $\{\Omega^j\}_{j=1}^{k}$ are further reinitialized after each loop through $j=1,\ldots,k$.
\end{itemize}

\begin{figure}[t!]
\centering
\rotatebox{90}{\parbox{2mm}{\textbf{MNIST}}}
\begin{subfigure}{.47\columnwidth}
  \centering
  \includegraphics[width=.99\linewidth]{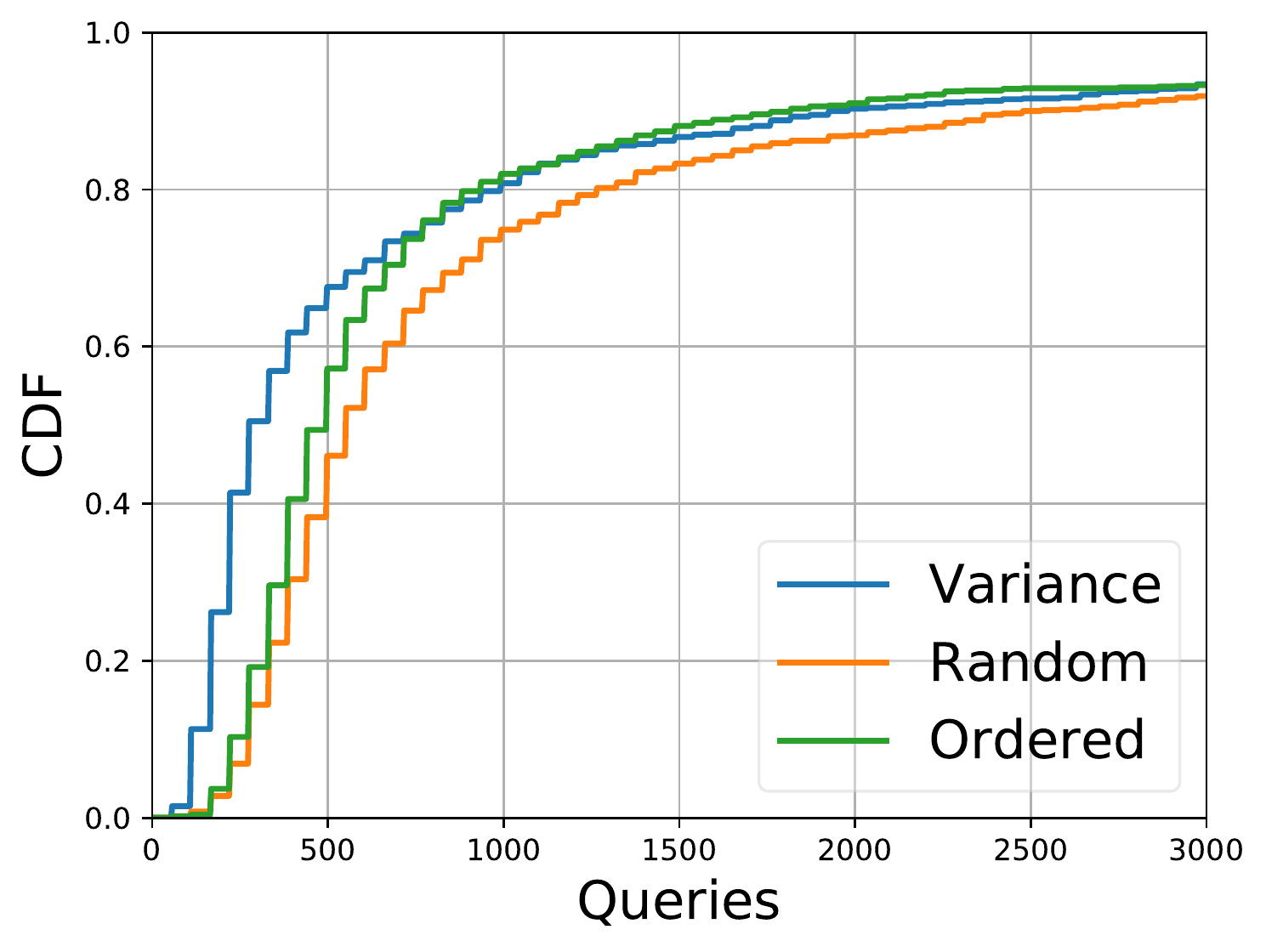}
  \caption{$\varepsilon_\infty = 0.4$}
\end{subfigure}%
\begin{subfigure}{.47\columnwidth}
  \centering
  \includegraphics[width=.99\linewidth]{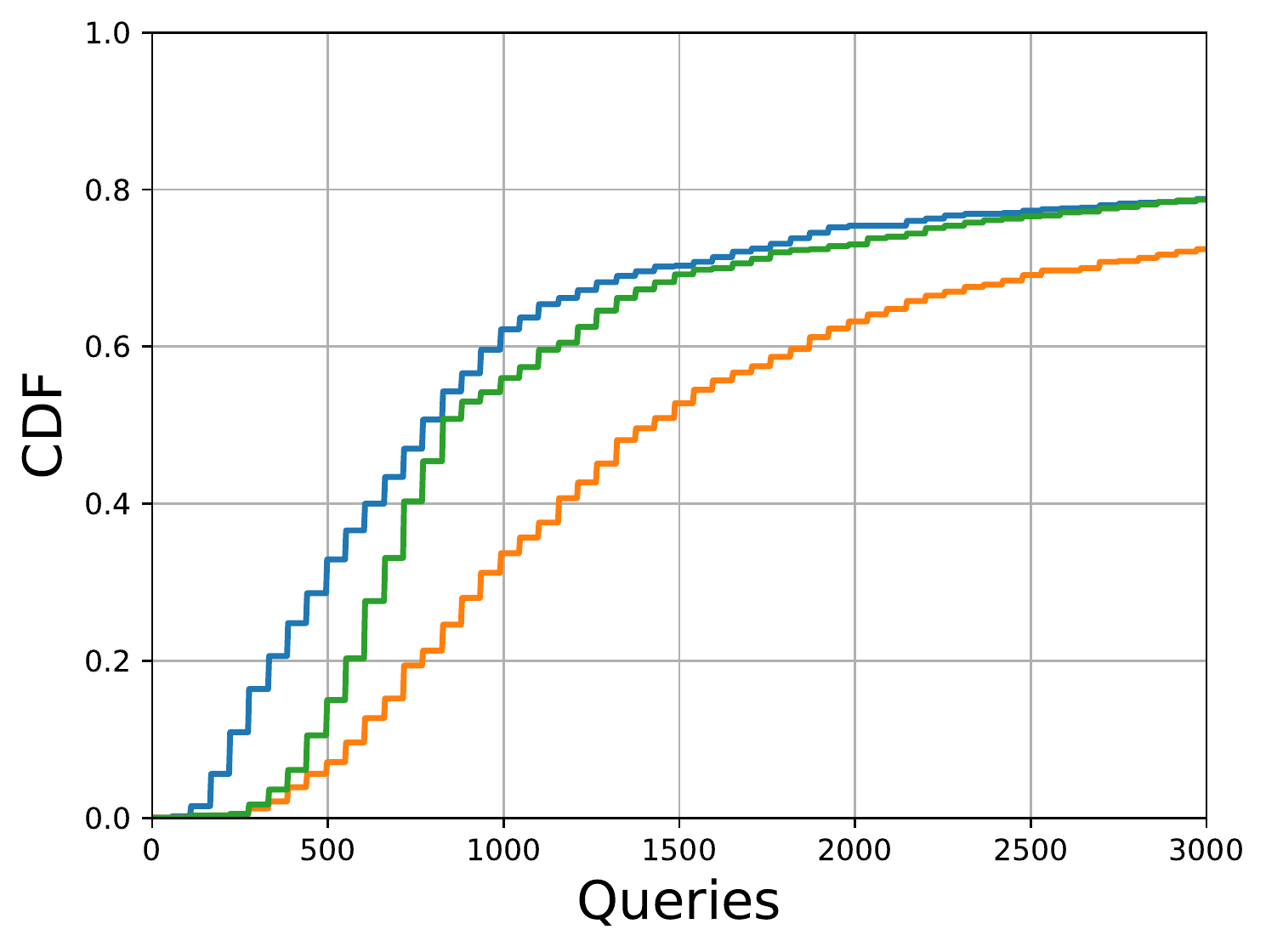}
  \caption{$\varepsilon_\infty = 0.2$}
\end{subfigure}\\
\rotatebox{90}{\parbox{2mm}{\textbf{CIFAR10}}}
\begin{subfigure}{.47\columnwidth}
  \centering
  \includegraphics[width=.99\linewidth]{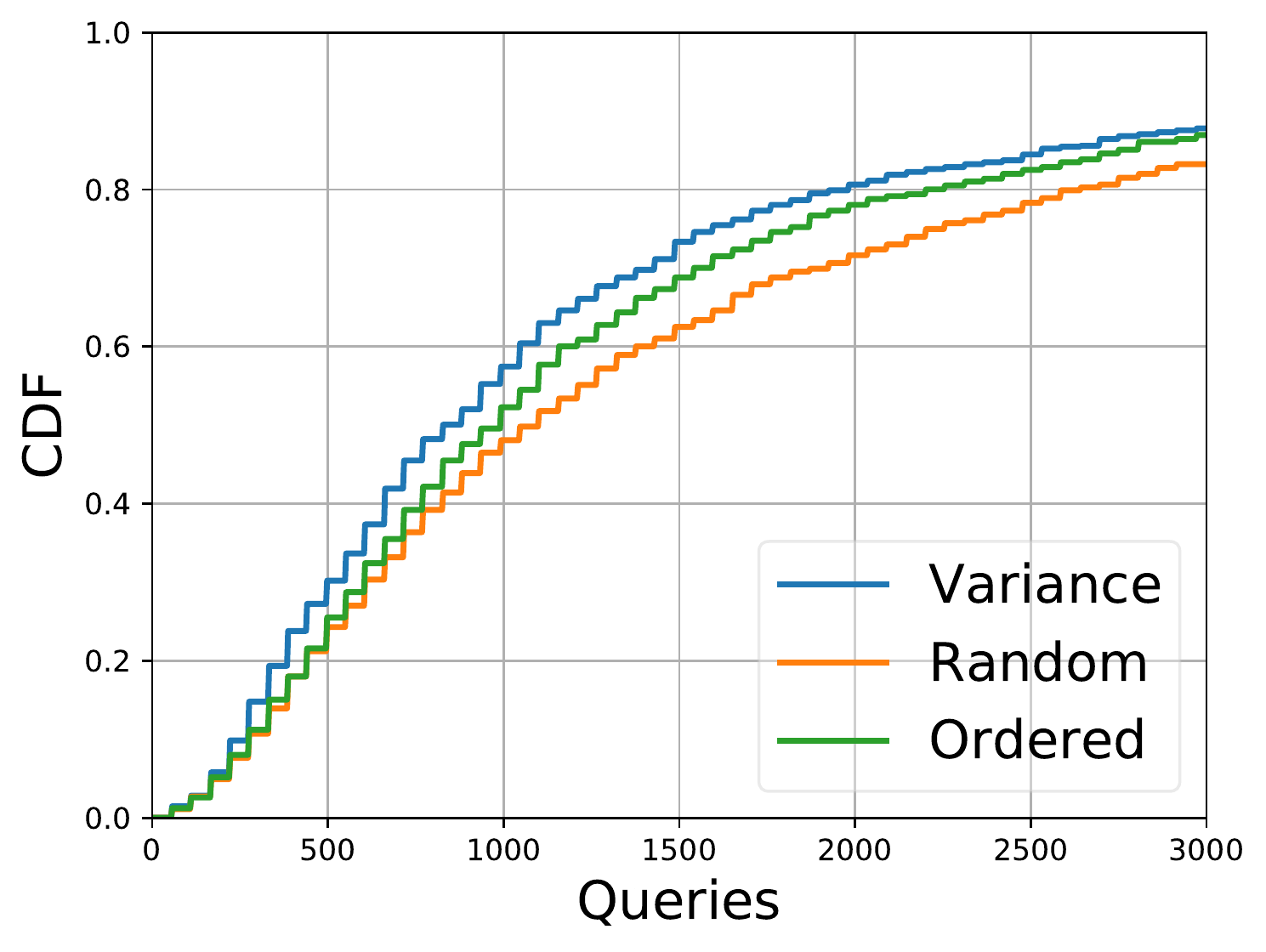}
  \caption{$\varepsilon_\infty = 0.1$}
\end{subfigure}%
\begin{subfigure}{.47\columnwidth}
  \centering
  \includegraphics[width=.99\linewidth]{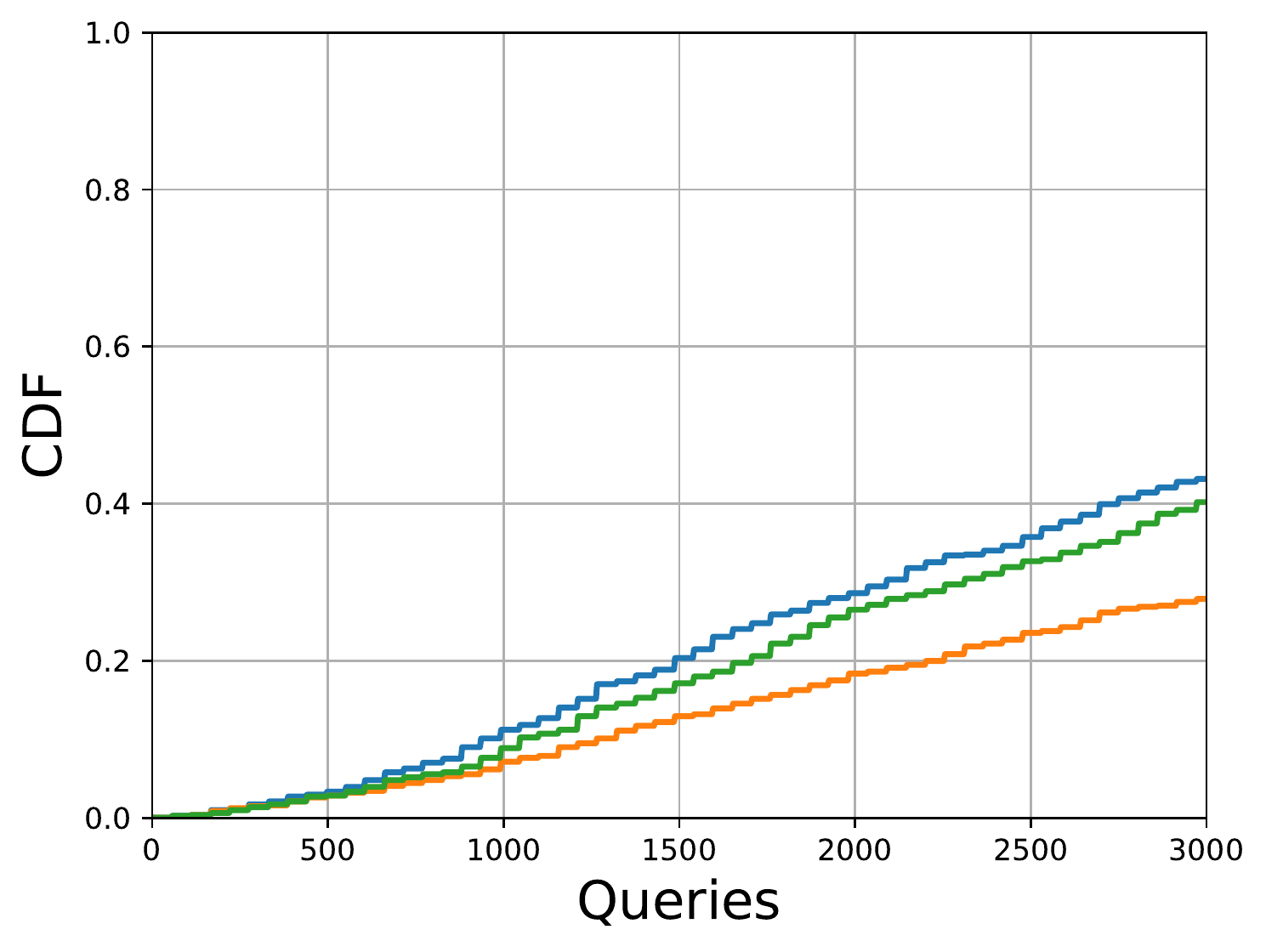}
  \caption{$\varepsilon_\infty = 0.02$}
\end{subfigure}%
\caption{Cumulative distribution function of successfully perturbed images as a function of number of queries by the BOBYQA based algorithm attacking DNNs trained on the MNIST and the CIFAR10 datasets. In each image the effectiveness of different sub-sampling methods in generating a successful adversarial example is shown for different values of maximum perturbation energies $\varepsilon_\infty$. See \cite{BOBYQA_Neurips} for details about experimental setup.}
\label{fig:subsamp}
\end{figure}

The sub-sampling of the domain affects the efficiency with which an algorithm successfully finds an adversarial example. For instance, in Figure \ref{fig:subsamp} we compare how these different sub-sampling techniques affect the BOBYQA based algorithm when generating adversarial example for the MNIST and CIFAR10 dataset.  It can be observed that variance sampling consistently has a higher success rate cumulative distribution function as compared with random and ordered sampling. This suggest that pixels belonging to high-contrast regions are more influential than the ones in low-contrast ones, and hence variance sampling is the preferable ordering.

To simplify the notation in the following section, the optimization variable is considered to be $\bm{\eta}^j=\bm{\Omega}^j \tilde{\bm{\eta}}^j$ where $\tilde{\bm{\eta}}^j \in \mathbb{R}^b$ and $\bm{\Omega}^j\in \mathbb{R}^{n \times b} $ is such that $[\bm{\Omega}^j]_{pq}$ is one if the $q$th element of $\Omega^j$ is $p$, zero otherwise. The implementation of variance sampling method at iteration $j$ in a domain of dimension $n_\ell$ is summarized in Algorithm 1.

\begin{algorithm}[t]
   \caption{GENERATE\_SAMPLING\_MATRIX($\hat{\textbf{X}}$,${n_\ell}$,b,j)}
   \label{alg:sampling_matrix}
\begin{algorithmic}[1]
   \STATE $\bm{\Omega}\leftarrow\textbf{0}\in\mathbb{R}^{{n_\ell}\times b}$
   \STATE $\textbf{v} \leftarrow$ argsort Var($\hat{\textbf{X}}$) \textit{\# Var defines the variance in intensity around a pixel}.
   \FOR{$i=1,\hdots,b$}
   \STATE $\bm{\Omega}(\textbf{v}[i + j\times b],[i])=1$.
   \ENDFOR
   \STATE Return $\bm{\Omega}$.
\end{algorithmic}
\end{algorithm}

\begin{figure}[t!]
\centering
\includegraphics[scale=.5]{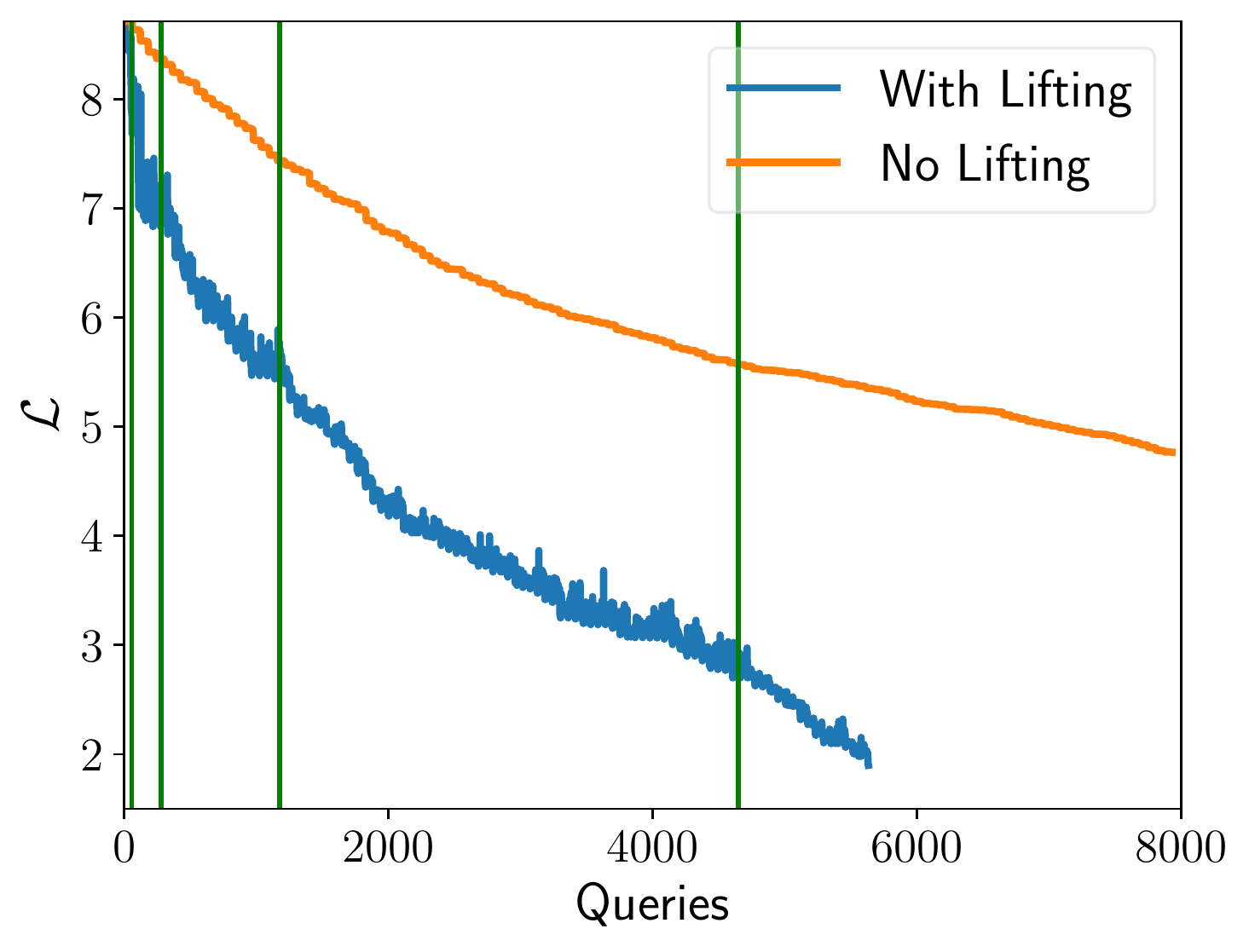}
\caption{Impact of hierarchical lifting approach on Loss function \eqref{eq:loss} as a function of the number of queries to Inception-v3 net trained on ImageNet dataset to find the adversarial example for a single image with the BOBYQA based method. The green vertical lines correspond to changes of hierarchical level, which entail an increase in the dimension of the optimization space.}
\label{fig:resize2}
\end{figure}

\subsection{Hierarchical Lifting}
\label{sec:hierarchical}

\begin{figure}[t!]
\centering
\includegraphics[scale=.4]{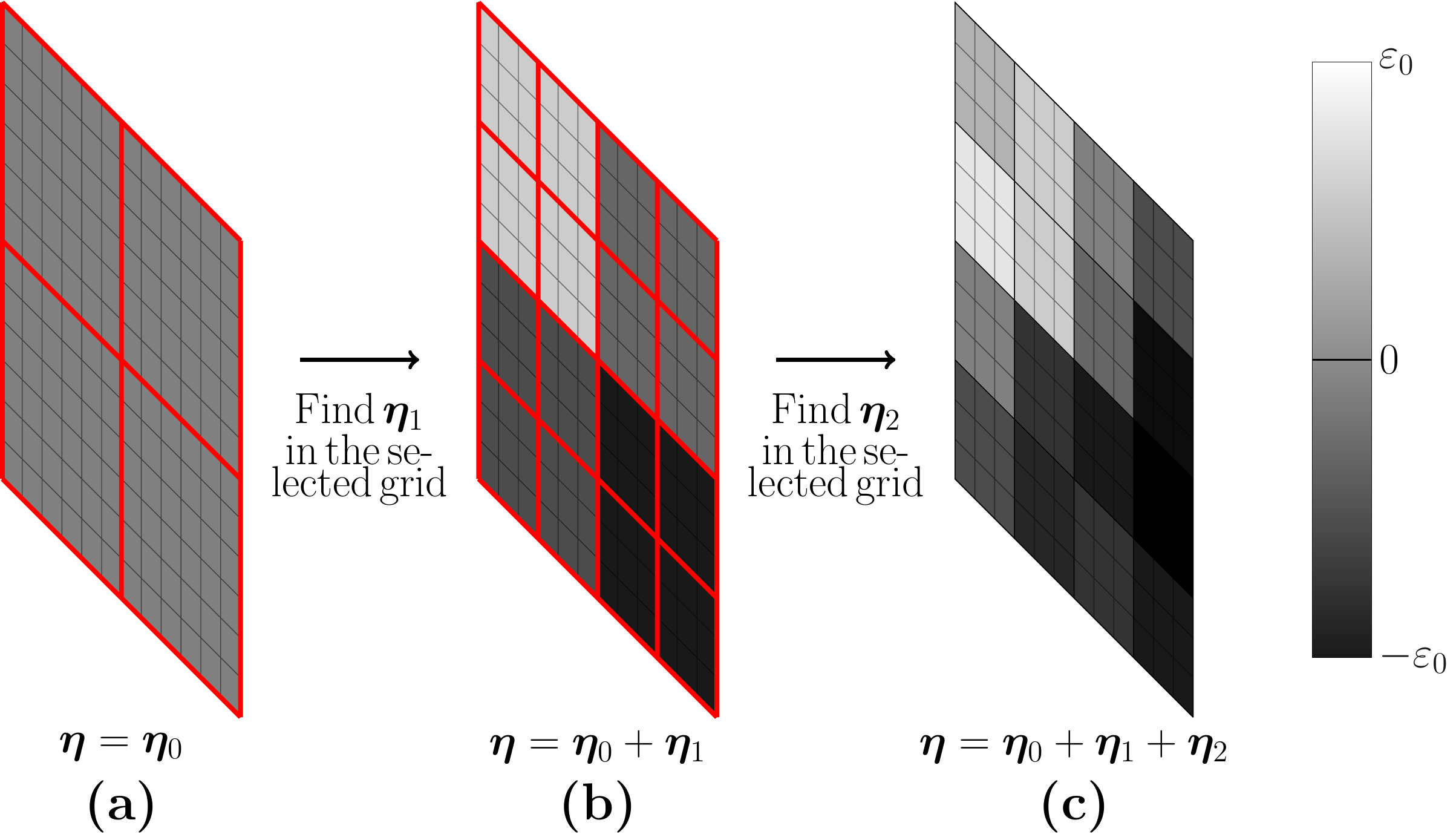}
\caption{Example of how the perturbation $\bm{\eta}$ is generated in a hierarchical lifting method with $n_1=4$ and $n_2=16$ on an image in $\mathbb{R}^{12\times12}$. In (a) the perturbation is $\bm{\eta} = \bm{\eta}_0$ and the boxes generated via the grid of dimension $n_1$ are highlighted in red. Once the optimal perturbation $\bm{\eta}_1$ is found, the perturbation is updated in (b) and the image is further divided with a grid with $n_2$ blocks. The final solution obtained after optimization is shown in (c).}
\label{fig:Projections}
\end{figure}

Authors of ZOO attack~\cite{chen} demonstrated that fewer queries are required to find adversarial example when pixels are considered in clusters, and not independently. This lead to the hierarchical lifting approach where one optimizes over increasingly higher dimensional spaces at each step, referred here as level $\ell$; Figure \ref{fig:resize2} shows how effective this approach is when implementing the BOBYQA based algorithm. These low dimensional spaces are lifted to the image space via a linear lifting, where at each level $\ell$ a linear lifting $\textbf{D}^\ell:\mathbb{R}^{n_\ell} \rightarrow \mathbb{R}^n$ is considered and a perturbation $\hat{\bm{\eta}}_\ell\in \mathbb{R}^{n_\ell}$ is found to be added to the full perturbation $\bm{\eta}$, according to
\begin{equation}
    \bm{\eta} = \sum_{j=0}^\ell \bm{\eta}_j= \sum_{j=0}^\ell \textbf{D}^j\hat{\bm{\eta}}_j.
\end{equation}
Here $\bm{\eta}_0$ is initialized as $\underline{\textbf{0}}$ and the perturbations $\bm{\eta}_j$ of the previous layers are considered as fixed. An example of how this works is illustrated in Figure \ref{fig:Projections}.

\begin{figure}
    \centering
 \begin{tabular}{c}
     \rotatebox{90}{\hspace{0.6cm}\large \textbf{ Random Lifting }}
\includegraphics[width=.85\linewidth]{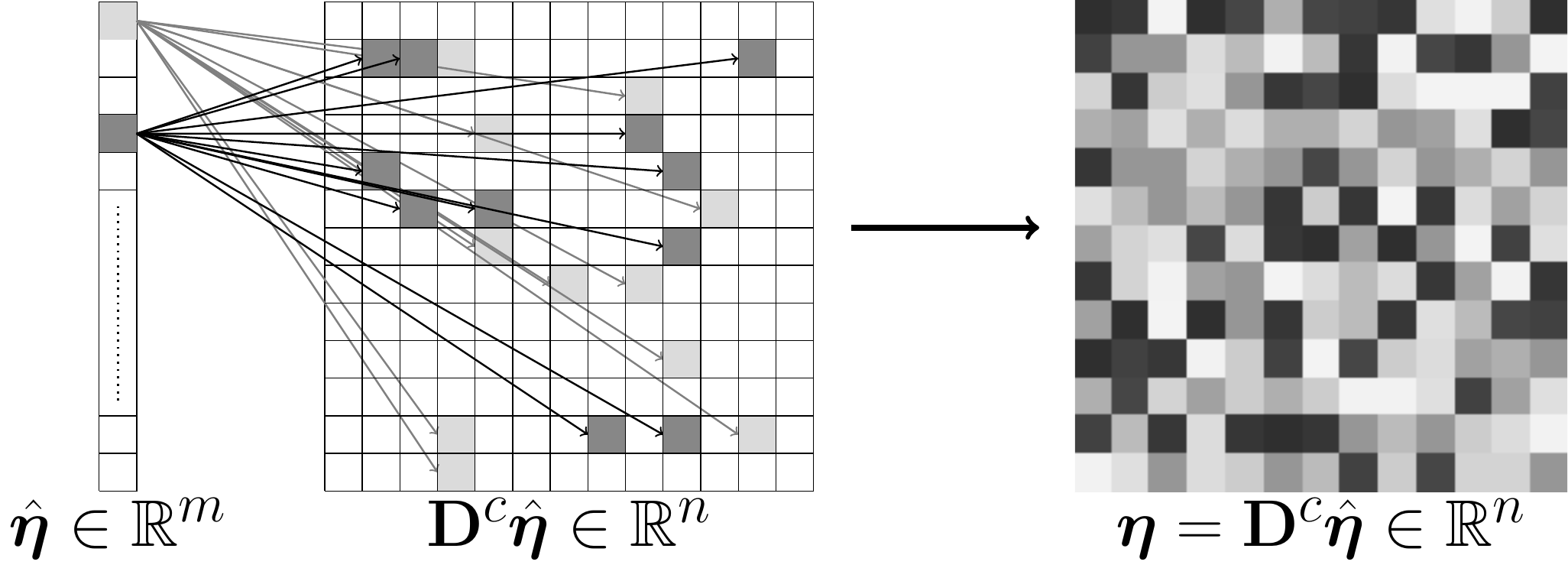}\\ 
\\ (a)\\
\\
\rotatebox{90}{\hspace{1.cm}\large \textbf{ Block Lifting }}
\includegraphics[width=.85\linewidth]{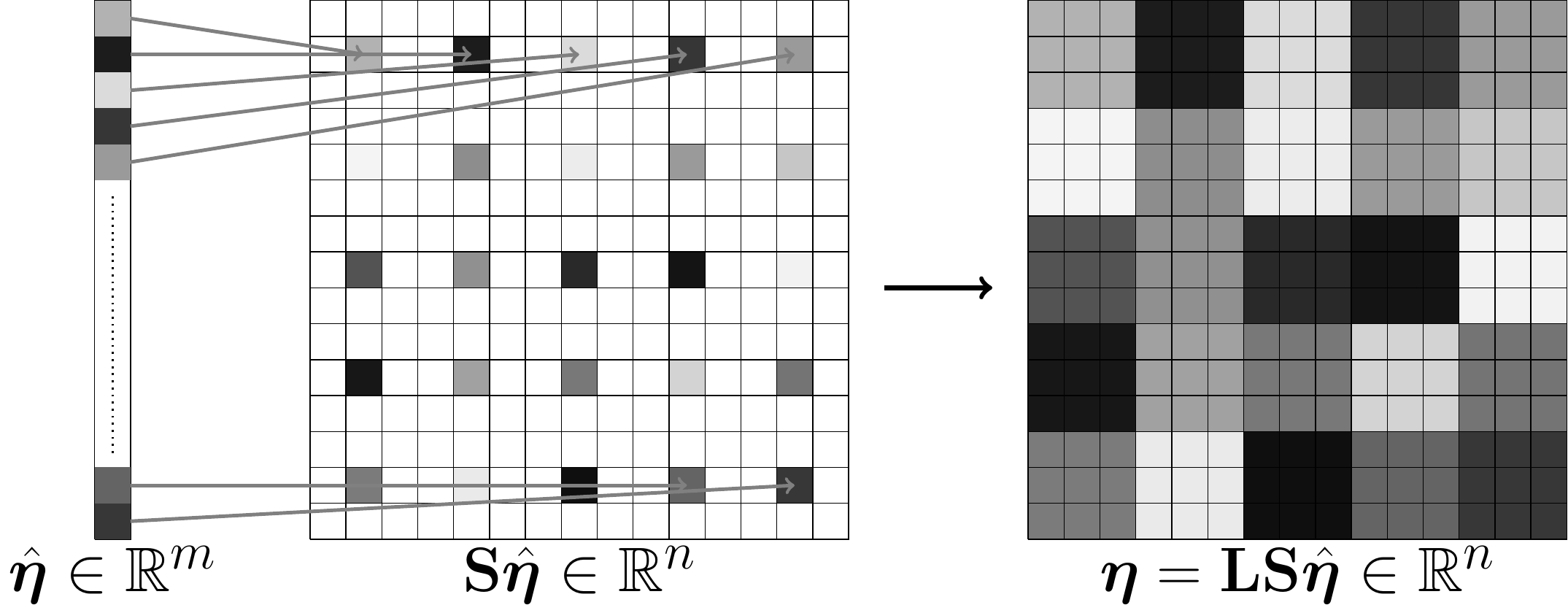}\\
(b)
 \end{tabular}
    \caption{ Examples for (a) random and (b) block liftings. In the random case each pixel in the perturbation is associated to just one element of $\hat{\bm{\eta}}_\ell$. Block lifting uses a piece-wise constant interpolation $\textbf{L}$ over a coarse grid $\textbf{S}\hat{\bm{\eta}}_\ell$ and each block is associated uniquely to one of the variables in $\hat{\bm{\eta}}_\ell$. In both cases, the lifting $\textbf{D}$ is such that each element $\textbf{D}_{ij}$ is either 1 or 0. 
}    \label{fig:Lifting}
\end{figure}

All the methods considered in this work rely on ideas which can be interpreted through this approach. The algorithms that we consider in this work rely on two kinds of linear lifting $\textbf{D}^\ell$ differentiated by the way each scalar in $\hat{\bm{\eta}}$ is associated to a set of pixels in the original image domain $\mathbb{R}^n$; namely the random and the block liftings. The former relates a random set of pixels of the original image to each hyper-variable; this forces the perturbation to be of high-frequency nature, as illustrated in Figure \ref{fig:Lifting}(a), which several articles indicate as being the most effective \cite{guo2018,gopalakrishnan2018toward,sharma2019effectiveness}. The GenAttack and Frank-Wolfe algorithms use a variation of this kind of lifting.  The latter instead is based on interpolation operations; a sorting matrix  $\textbf{S}^\ell:\mathbb{R}^{n_\ell}\rightarrow\mathbb{R}^n$ is applied such that every index of $\hat{\bm{\eta}}_\ell$ is uniquely associated to a node of a coarse grid masked over the original image. Afterwards, an interpolation $\textbf{L}^\ell:\mathbb{R}^n\rightarrow\mathbb{R}^n$ is implemented over the values in the coarse grid, i.e. $\bm{\eta}_\ell = \textbf{L}^\ell \textbf{S}^\ell\hat{\bm{\eta}}_\ell = \textbf{D}^\ell\hat{\bm{\eta}}_\ell$. Both Square and Parsimonious algorithms implement hierarchical lifting with the piece-wise constant interpolation, here referred to as block lifting. At the lower levels the interpolation lifting generates low frequency perturbations, as illustrated in Figure \ref{fig:Lifting}(b).

Since $n_\ell$ may still be very high, for each level $\ell$ domain sub-sampling is also applied considering $\hat{\bm{\eta}}_\ell = \sum_{j=0}^k \tilde{\bm{\eta}}_\ell^j$. In the piece-wise constant case with variance sampling, the blocks are ordered according to the variance of mean intensity among neighboring blocks, in contrast to the variance within each block as suggested in \cite{chen}. Consequently, at each level the adversarial example is found by solving the following iterative problem
\begin{align}
    \min_{\tilde{\bm{\eta}}_\ell^j} \ \ \ & \mathcal{L}\left(\textbf{X} + \bar{\bm{\eta}},   \textbf{D}^\ell\bm{\Omega}^k\tilde{\bm{\eta}}_\ell^j\right) \;\label{eq:opt_sub_samp_fin}\\ \text{s.t.}\quad &\;\left\|\bar{\bm{\eta}} +  \textbf{D}^\ell\bm{\Omega}^k\tilde{\bm{\eta}}_\ell^j\right\|_\infty \leq \varepsilon_\infty \nonumber\\
    & \left[\textbf{X} + \bar{\bm{\eta}} +  \textbf{D}^\ell\bm{\Omega}^k\tilde{\bm{\eta}}_\ell^j\right]_r \geq l \quad\quad\quad\;\; \forall r\in\{1,...,n\} \nonumber\\
    & \left[\textbf{X} + \bar{\bm{\eta}} +  \textbf{D}^\ell\bm{\Omega}^k\tilde{\bm{\eta}}_\ell^j\right]_r \leq u \quad\quad\quad \forall r\in\{1,...,n\},\nonumber \label{eq:opt_prob_proj_end}
\end{align}
where $\bar{\bm{\eta}} = \sum_{i=0}^{\ell-1}\bm{\eta}_i + \textbf{D}^\ell \sum_{m\neq j} \hat{\bm{\eta}}_\ell^m$. Algorithm \ref{alg:ALG_LIFTING} gives an implementation of the block lifting matrix when in the grid has dimension $n_\ell$.

\begin{algorithm}[t]
   \caption{GENERATE\_LIFTING(${n_\ell}$,n)}
   \label{alg:ALG_LIFTING}
\begin{algorithmic}[1]
   \STATE $\textbf{D}\leftarrow\textbf{0}\in\mathbb{R}^{n\times n_\ell}$
   \FOR{$i=1,\hdots,{n_\ell}$}
   \STATE Generate set of pixels $S$ that are in the block associated to the $i$-th element of the ${n_\ell}$ dimensional super-grid.
   \FOR{$j \in S$}
   \STATE $\textbf{D}(i,j)=1$.
   \ENDFOR
   \ENDFOR
   \STATE Return $\textbf{D}$.
\end{algorithmic}
\end{algorithm}

\section{Comparison of Derivative Free Methods}
\label{sec:exp}

In this section, we compare algorithms based on a selection of state-of-the-art DFO methods.  In particular we consider BOBYQA based algorithm~\cite{BOBYQA_Neurips}, GenAttack algorithm~\cite{Alzantot}, Parsimonious algorithm~\cite{COMBI}, Square algorithm~\cite{andriushchenko2019square} and Frank-Wolfe algorithm~\cite{chen2020frank} in the following two frameworks:
\begin{itemize}
    \item Section \ref{sec:knwon_experiments} considers the canonical setup for black-box adversarial attacks on which the considered algorithms have been tuned in their respective articles. Specifically, we consider attacks on networks trained adversarially or not on CIFAR10 and ImageNet, two popular datasets in the literature, and with no further defense implemented.
    
    \item Section \ref{sec:unkonown_exp}  considers a setup that simulates structural defenses on which the different algorithms were not tuned. We limit the perturbation to a fixed number of pixels with high variance in intensity considering attacks on a network non-adversarially trained on the CIFAR10 dataset.
\end{itemize}

The performance of all algorithms is measured in terms of the distribution of queries needed to successfully find adversaries to identical networks given a fixed $\ell^\infty$ perturbation constraint and the same input images.

\subsection{Parameter Setup for Algorithms}

The experiments use publicly available implementations for the GenAttack \cite{Alzantot}, Parsimonious \cite{COMBI}, Square \cite{andriushchenko2019square}, and Frank-Wolfe \cite{chen2020frank} algorithms\footnote{GenAttack: \url{https://github.com/nesl/adversarial_genattack}\\ Parsimonious algorithm: \url{https://github.com/snu-mllab/parsimonious-blackbox-attack} \\ Square algorithm: \url{https://github.com/max-andr/square-attack} \\ Frank-Wolfe algorithm \url{https://github.com/uclaml/Frank-Wolfe-AdvML}} using the same hyper-parameter setting and hierarchical lifting approach as suggested by the respective authors. 

For the BOBYQA based algorithm~\cite{BOBYQA_Neurips}, from Figure~\ref{fig:subsamp} we observed that the loss function is influenced the most by the pixels in high-contrast areas. Hence, we first apply the variance sub-sampling method followed by block lifting as described in Section~\ref{sec:hierarchical}\footnote{The choice for this kind of lifting was driven by preliminary experiments in which we considered also a grid method with linear interpolation and a random lifting method as well. It is possible to run the analysis thanks to the code in $^3$}. Here, we consider an initial domain of dimension $n_1 = 2 \times 2 \times 3$, and double the refinement of the grid at each layer, i.e. $n_{\ell+1} = 4n_\ell$. Moreover, we observe for \eqref{eq:update_model}, the choice of a linear model to approximate the loss function works best, and we consequently consider the linear approximation in this paper; i.e., $\textbf{M}=\textbf{0}$ and $q=n+1$ at all iterations, see \cite{BOBYQA_Neurips}. The BOBYQA based algorithm is summarized in Algorithm \ref{alg:ALG} and a Python implementation of the proposed algorithm based on BOBYQA package from \cite{cartis} is available on Github\footnote{\url{https://github.com/giughi/An-Empirical-Study-of-DFO-Algorithms-for-Targeted-Black-Box-Attacks-in-DNNs}}.

\begin{algorithm}[t]
   \caption{BOBYQA Based Algorithm}
   \label{alg:ALG}
\begin{algorithmic}[1]
   \STATE {\bfseries Input:} Image $\textbf{X}\in \mathbb{R}^n$, target label $t$, maximum perturbation $\varepsilon_\infty$, Neural Net $F$, initial hierarchical level grid dimensions $m$, maximum number of queries $n^{max}$, batch sampling size $b$, and maximum number $\kappa$ of queries that we are allowed to do for each batch.
   \STATE \textbf{Initialize} $\bm{\eta} \leftarrow \underline{0} \in \mathbb{R}^n$, $n_{eval} = 0$, $\ell=1$, $n_\ell=12$.
   
   \WHILE{$\argmax F(\textbf{X}+\bm{\eta})\neq t$ and $n_{eval}<n^{max}$}
   \STATE \textit{ \# Compute the number of sub samplings necessary to cover the whole domain}
   \STATE  $num_{sub}=n/({n_\ell}*b)$ \quad \quad \quad  
   \STATE \textit{\# Generate the lifting matrix }
   \STATE $\textbf{D}_\ell$ = GENERATE\_LIFTING(${n_\ell},n$) \quad \quad 
   \STATE \textit{\# Minimize on all the sampled sub-domains}
   \FOR{$j = 1, \hdots, num_{sub}$}
   \STATE \textit{\# Compute the  matrix which selects $b$ dimensions of the $m$-dimensional domain.}
   \STATE $\bm{\Omega}_\ell^j$ = GENERATE\_SAMPLING\_MATRIX($\textbf{X}+\bm{\eta},{n_\ell},b,j$)
   \STATE \textit{\# Define the pixel-wise bounds for a perturbation over $\textbf{X} + \bm{\eta}$.}
   \STATE $\textbf{a} = \min\{ l - \bm{\eta} ,0\}$, \quad $\textbf{b} = \max \{u - \bm{\eta} ,0\}$
   \STATE \textit{\# Find $\hat{\bm{\eta}}_\ell^j$ by implementing the BOBYQA optimization to the problem (\ref{eq:opt_sub_samp_fin}).}
   \STATE $\hat{\bm{\eta}}_\ell^j$=BOBYQA($F,\textbf{X}, \bm{\eta}, \textbf{a},\textbf{b}, \textbf{D}_\ell, \bm{\Omega}_\ell^j,t$) \quad{\# Algorithm \ref{alg:ALG_BOBYQA}}
   \STATE \textit{\# Update the noise} 
   \STATE $\bm{\eta} += \textbf{D}_\ell\bm{\Omega}_\ell^j \hat{\bm{\eta}}_\ell^j$.
   \STATE $n_{eval}$ += $\kappa$.
   \ENDFOR
   \STATE $\ell+=1$, ${n_\ell}*=4$.
   \ENDWHILE
   \IF{$\argmax F(\textbf{X}+\bm{\eta})= t$}
   \STATE The perturbation is successful.
   \ELSIF{$n_{eval} > n^{max}$}
   \STATE The perturbation was not successful with $n^{max}$ iterations.
   \ENDIF
\end{algorithmic}
\end{algorithm}

\begin{algorithm}[t]
   \caption{BOBYQA($F$, $\textbf{X}$, $\bm{\eta}$, $\textbf{a}$, $\textbf{b}$, $\bm{\Omega}_\ell^j$, $\textbf{D}^\ell$, $t$, $\kappa$)}
   \label{alg:ALG_BOBYQA}
\begin{algorithmic}[1]
   \STATE Consider the restricted loss function $\mathcal{L}(\textbf{X} + \bm{\eta}, \textbf{D}^\ell\bm{\Omega}_\ell^j(\cdot)): \mathbb{R}^b \rightarrow \mathbb{R}$
   \STATE Build an initial model $m_0$ as in \eqref{eq:Model} of the loss function based on $b+1$ samples; the samples consist of the initial perturbation $\textbf{X} + \bm{\eta}$ and the $b$ perturbations obtained by considering changes along the canonical directions of $\textbf{x}$ in $\textbf{X} + \bm{\eta} + \textbf{D}^\ell\bm{\Omega}_\ell^j\textbf{x}$.
   \STATE Find minimizer $\textbf{x}$ of $m_o$ such that $\textbf{D}^\ell\bm{\Omega}_\ell^j\textbf{x}\in[\textbf{a},\textbf{b}]$.
   \FOR{$j = 1, \hdots, \kappa - b$}
   \STATE Add $\textbf{x}$ to the set of samples and get rid of the least informative one according to \cite{powellbobyqa}.
   \STATE Build the new model $m_j$ according to \eqref{eq:update_model}.
   \STATE Find minimizer $\textbf{x}$ of $m_j$ such that $\textbf{D}^\ell\bm{\Omega}_\ell^j\textbf{x}\in[\textbf{a},\textbf{b}]$.
   \ENDFOR
    \STATE Return $\textbf{x}$.
\end{algorithmic}
\end{algorithm}

\subsection{Dataset and Neural Network Specifications}
We performed experiments using the popular ResNet50 architecture~\cite{he2016deep} with two training scenarios; one with the unperturbed images, and one with the defense\footnote{These networks are available already trained at \url{https://github.com/MadryLab/robustness}} proposed in~\cite{robustness}. The number of experiments and the choice of the targets for each individual dataset is described below.

\begin{figure*}[t!]
    \centering
        \begin{center}
        \begin{subfigure}{.8\textwidth}
          \centering
          \includegraphics[width=.99\linewidth]{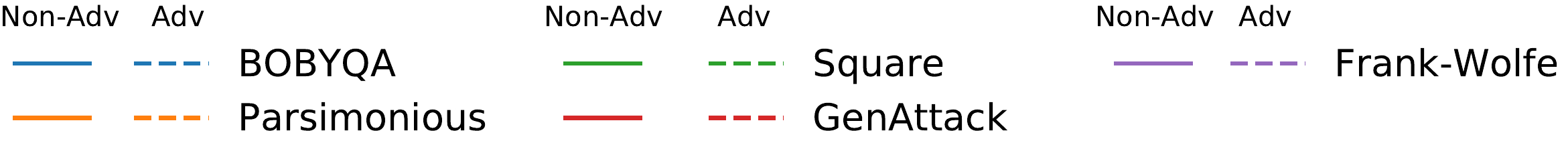}
        \end{subfigure}\\
        \vspace{0.2cm}
        \begin{subfigure}{.45\textwidth}
          \centering
          \includegraphics[trim={0.25cm 0.35cm 0.3cm 0.26cm},clip,width=.99\linewidth]{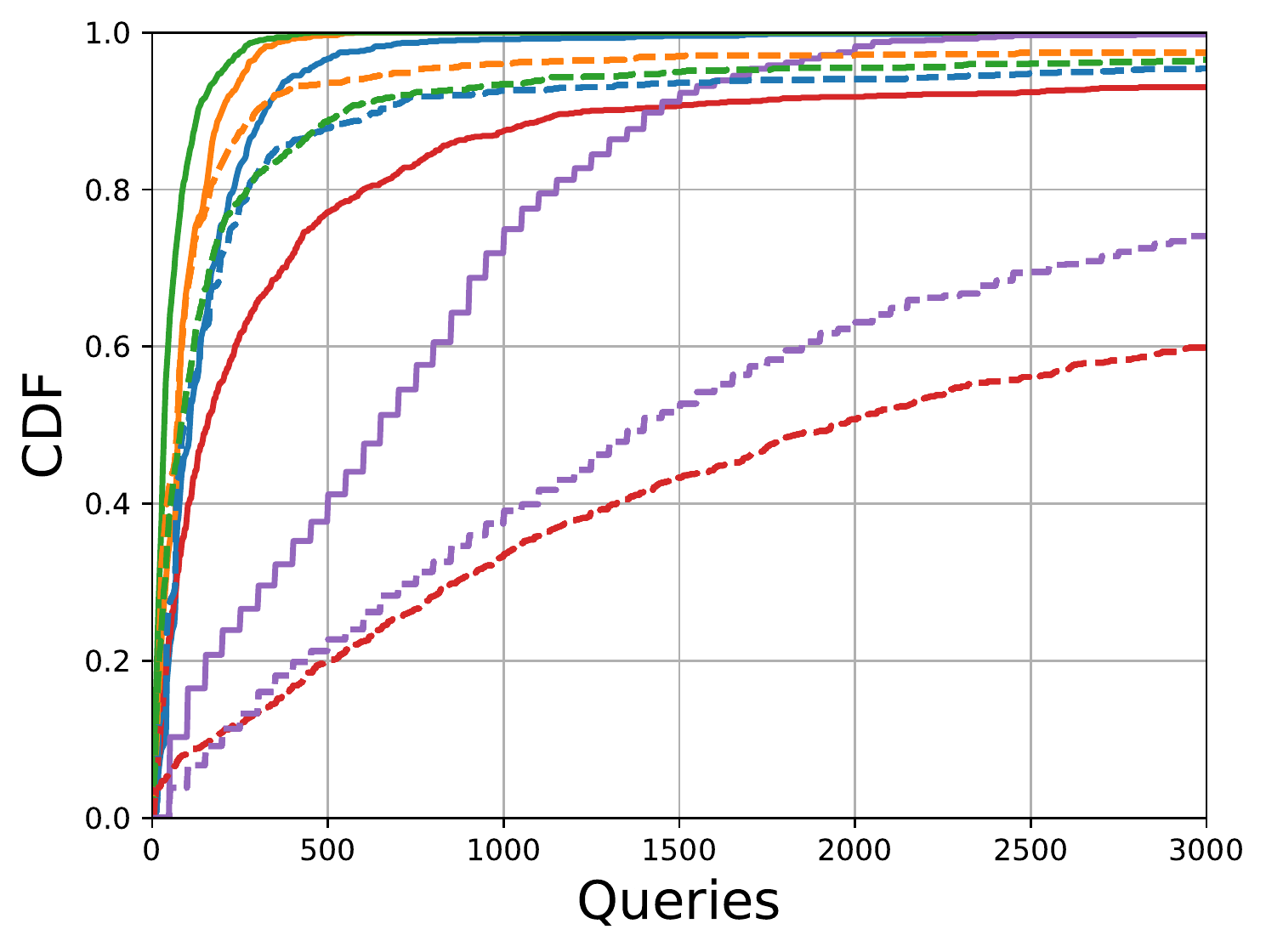}
          \caption{$\varepsilon_\infty=0.15$}
        \end{subfigure}%
        \begin{subfigure}{.45\textwidth}
          \centering
          \includegraphics[trim={0.25cm 0.35cm 0.3cm 0.26cm},clip,width=.99\linewidth]{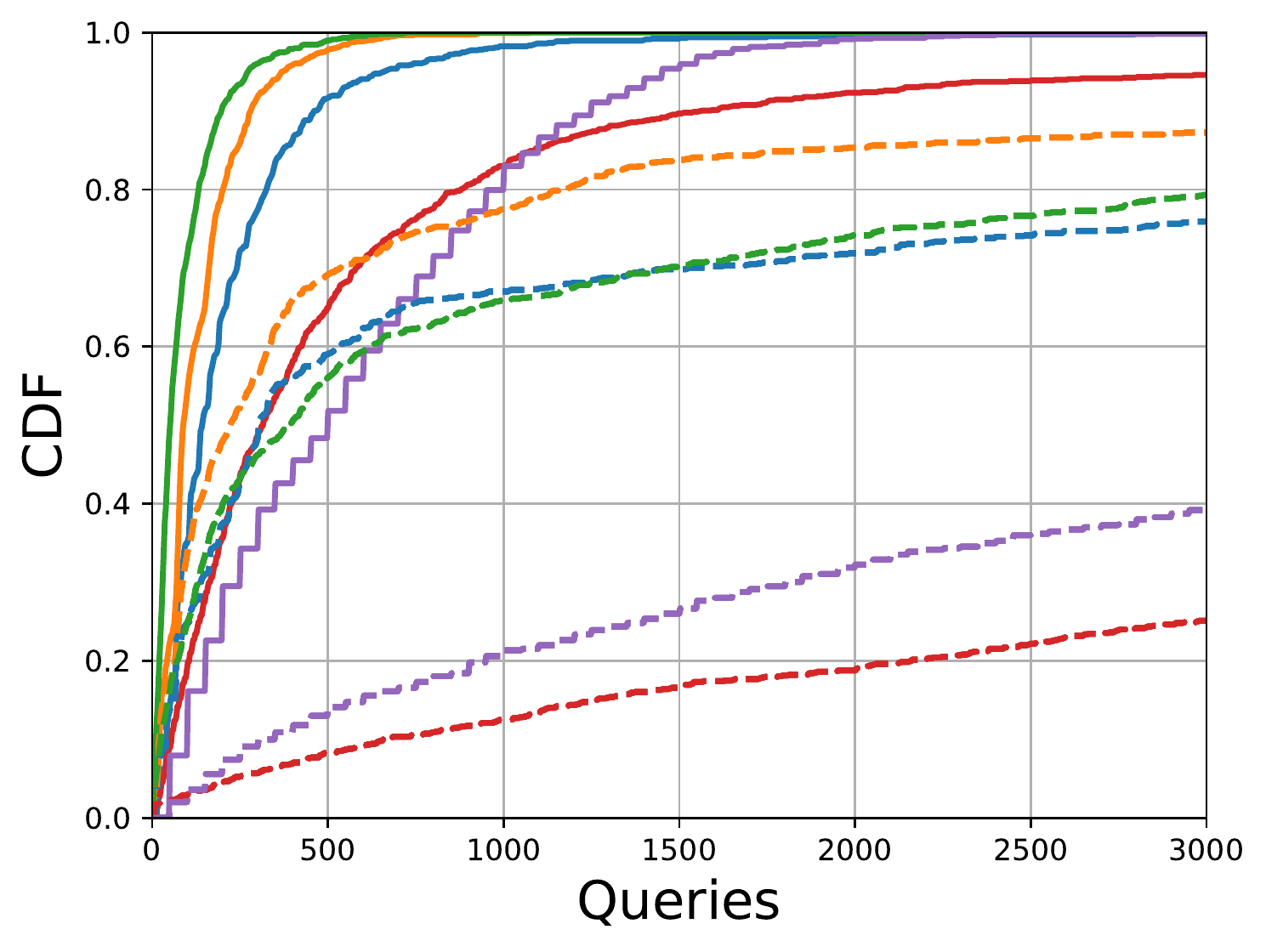}
          \caption{$\varepsilon_\infty=0.1$}
        \end{subfigure} \\
        \begin{subfigure}{.45\textwidth}
          \centering
          \includegraphics[trim={0.25cm 0.35cm 0.3cm 0.26cm},clip,width=.99\linewidth]{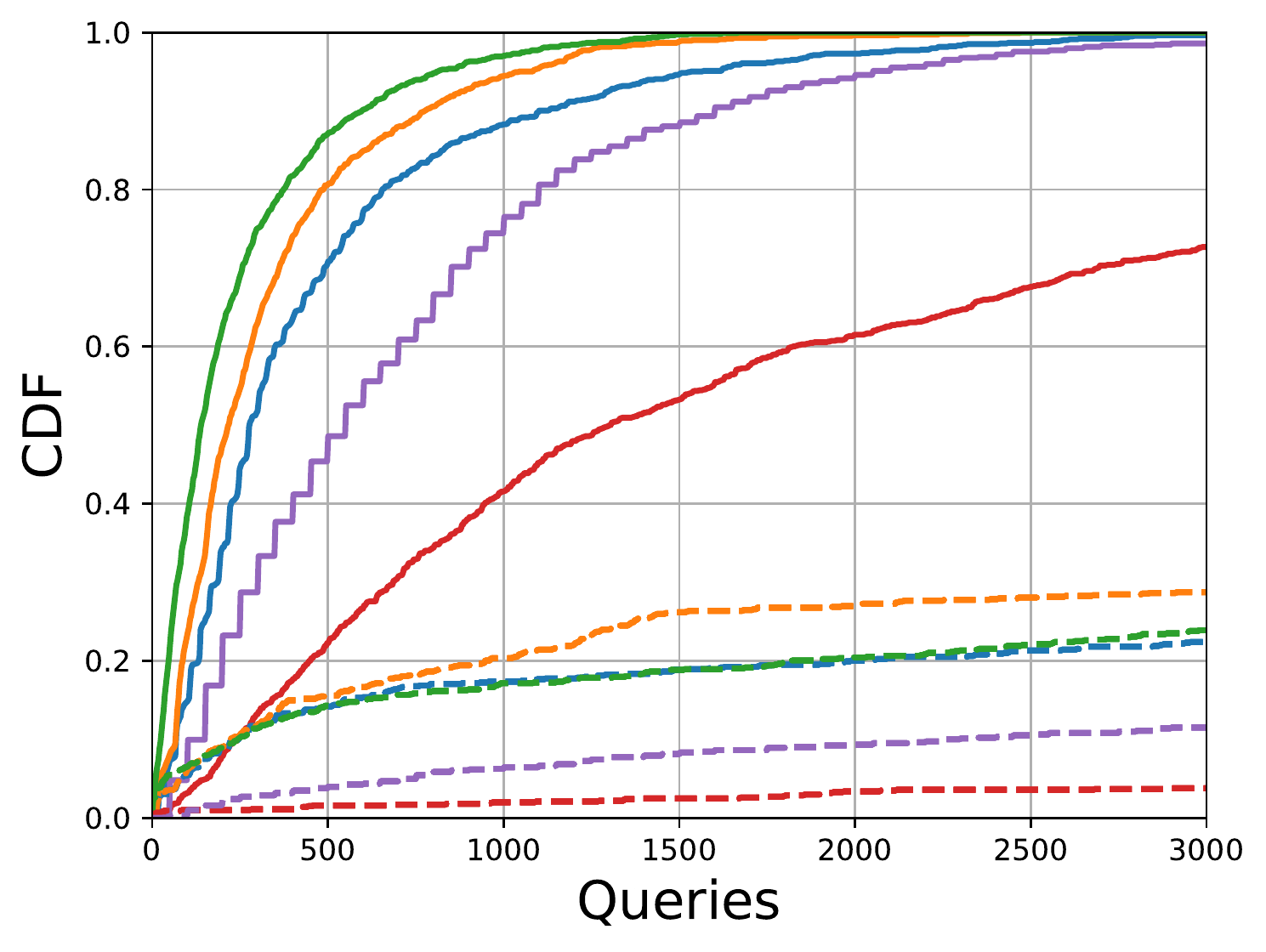}
          \caption{$\varepsilon_\infty=0.05$}
        \end{subfigure}%
        \begin{subfigure}{.45\textwidth}
          \centering
          \includegraphics[trim={0.25cm 0.35cm 0.3cm 0.26cm},clip,width=.99\linewidth]{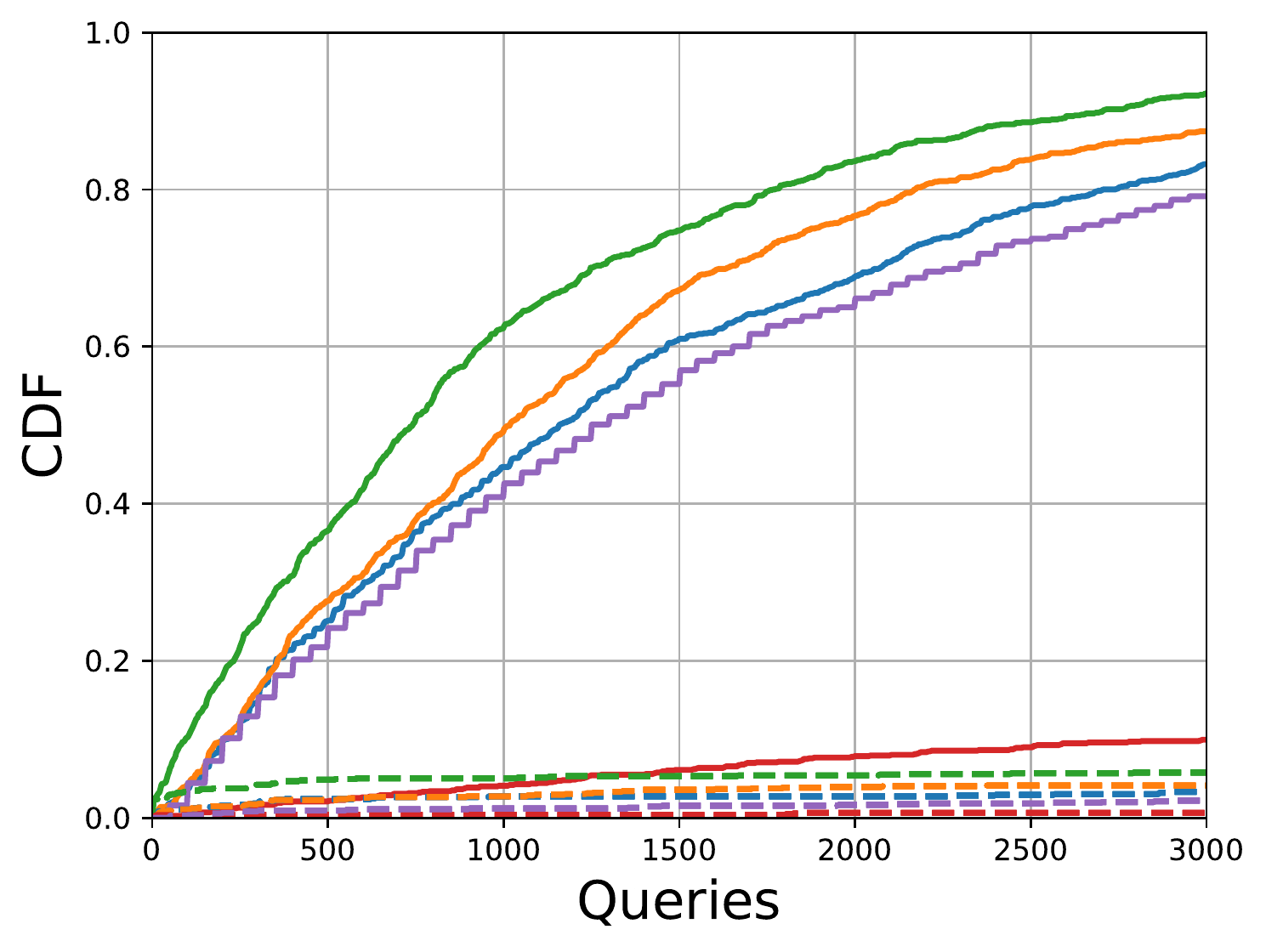}
          \caption{$\varepsilon_\infty=0.02$}
        \end{subfigure}\\
        \begin{subfigure}{.45\textwidth}
          \centering
          \includegraphics[trim={0.25cm 0.35cm 0.3cm 0.26cm},clip,width=.99\linewidth]{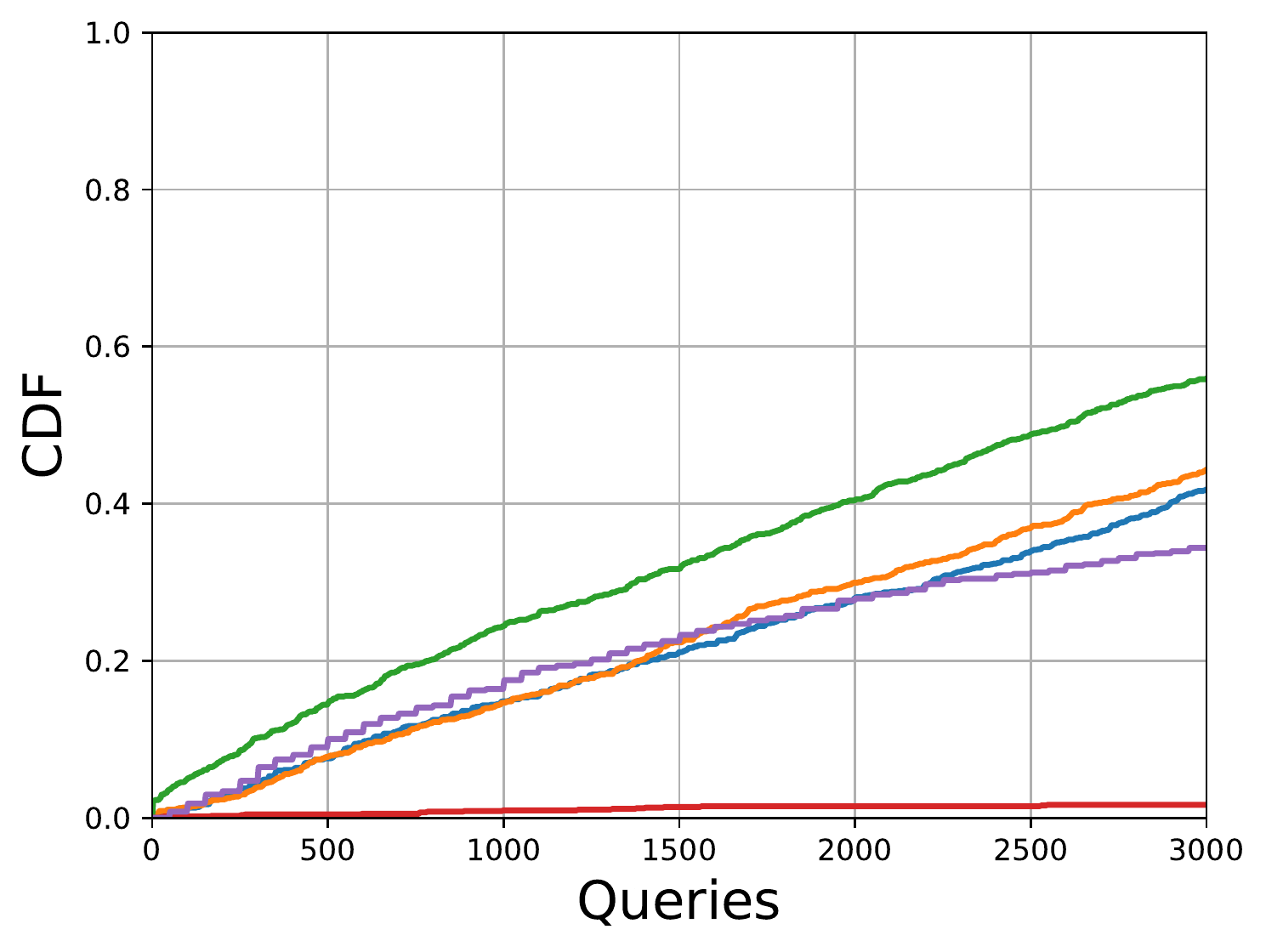}
          \caption{$\varepsilon_\infty=0.01$}
        \end{subfigure}%
        \begin{subfigure}{.45\textwidth}
          \centering
          \includegraphics[trim={0.25cm 0.35cm 0.3cm 0.26cm},clip,width=.99\linewidth]{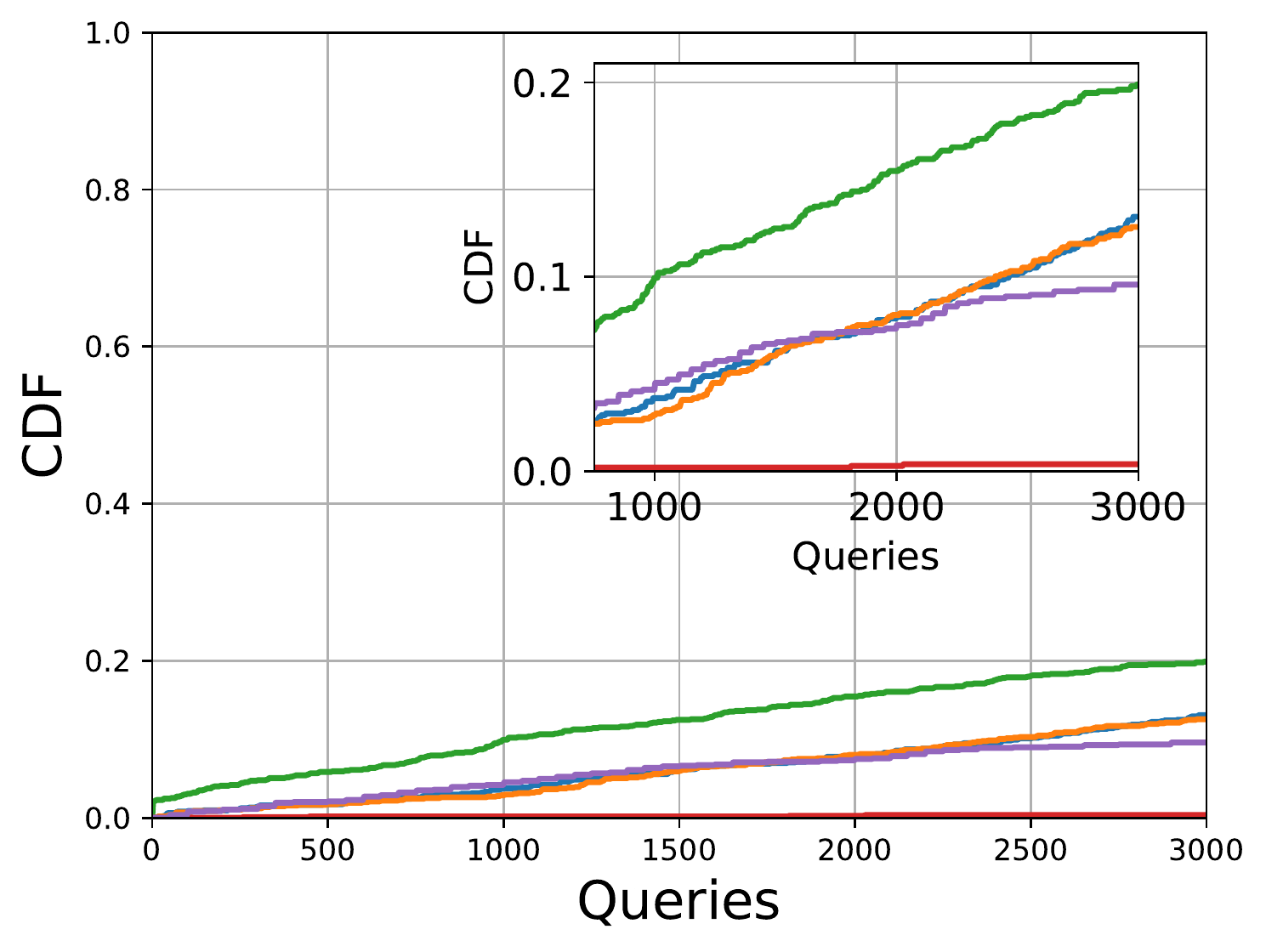}
          \caption{$\varepsilon_\infty=0.005$}
        \end{subfigure}
        \end{center}
        \caption{Cumulative fraction of test set images successfully misclassified with adversarial examples generated by GenAttack, Parsimonious, Square, Frank-Wolfe, and our BOBYQA based approaches for different maximum perturbation energies $\varepsilon_\infty$ and DNNs trained on the CIFAR10 dataset. In all results the solid and dashed lines denoted by `Non-Adv' and `Adv' corresponds to attacks on networks trained without or with the MadryLab defense strategy \cite{robustness} respectively.
        }
        \label{fig:CIFARcdf}
\end{figure*}

\begin{figure*}[t!]
    \centering
        \begin{subfigure}{.8\textwidth}
          \centering
          \includegraphics[width=.99\linewidth]{FIGURE_PYTORCH/ImageNet/Labels_True.pdf}
        \end{subfigure}\\
        \vspace{0.2cm}
        \begin{subfigure}{.45\textwidth}
          \centering
          \includegraphics[trim={0.25cm 0.35cm 0.3cm 0.26cm},clip,width=.99\linewidth]{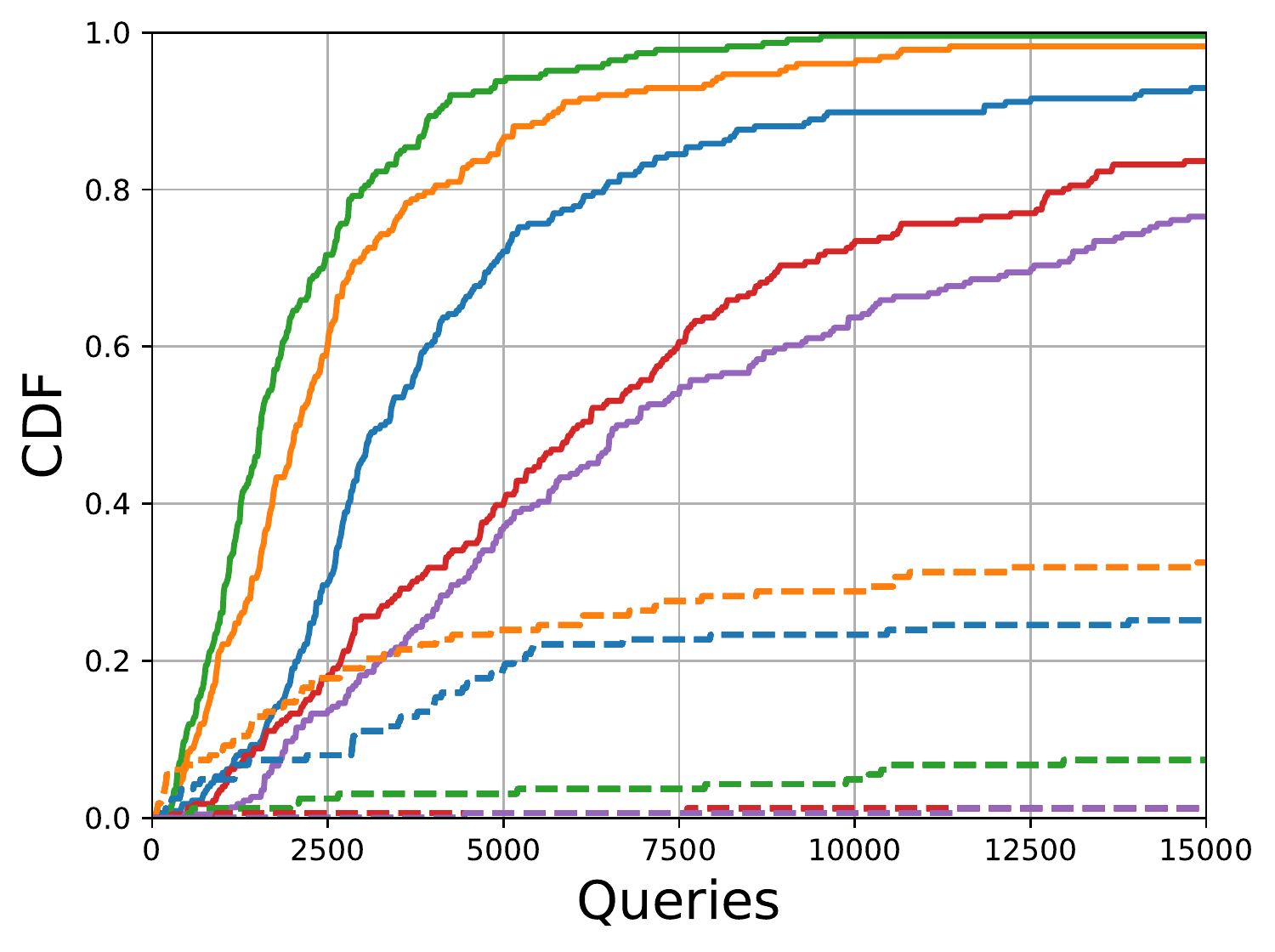}
          \caption{$\varepsilon_\infty=0.1$}
          \label{fig:imagenetsub1}
        \end{subfigure}%
        \begin{subfigure}{.45\textwidth}
          \centering
          \includegraphics[trim={0.25cm 0.35cm 0.3cm 0.26cm},clip,width=.99\linewidth]{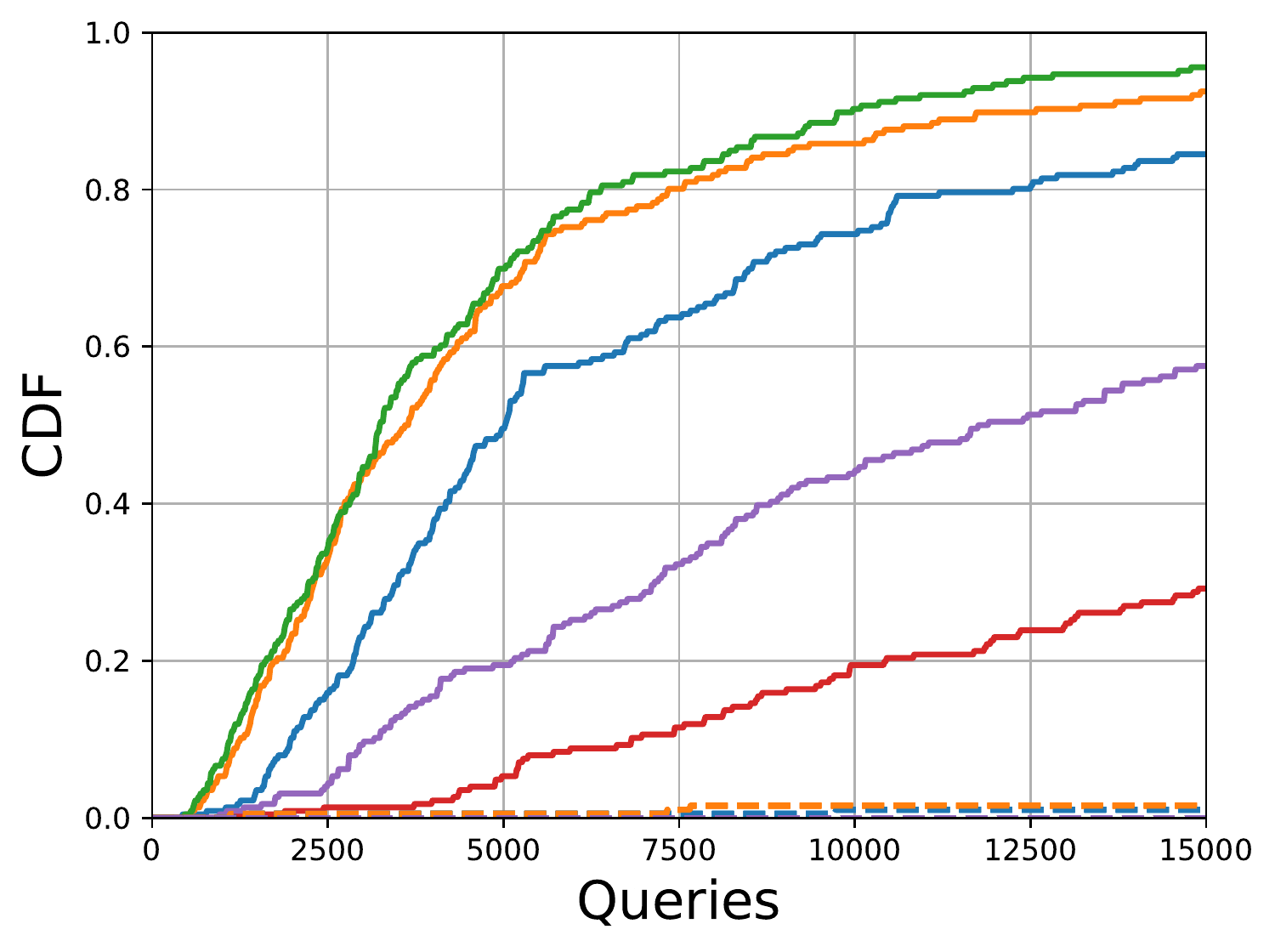}
          \caption{$\varepsilon_\infty=0.05$}
        \end{subfigure} \\
        \begin{subfigure}{.45\textwidth}
          \centering
          \includegraphics[trim={0.25cm 0.35cm 0.3cm 0.26cm},clip,width=.99\linewidth]{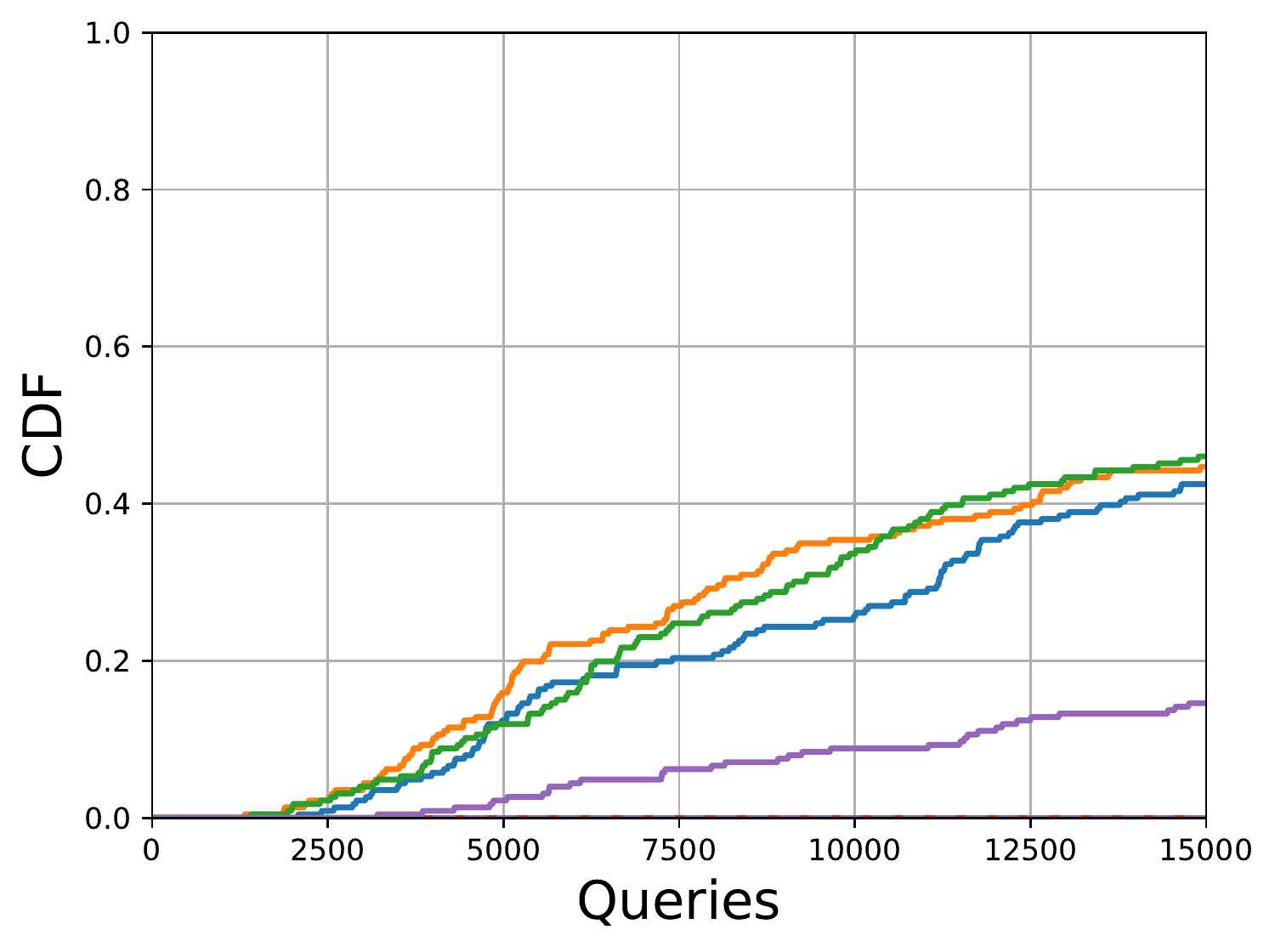}
          \caption{$\varepsilon_\infty=0.02$}
        \end{subfigure}%
        \begin{subfigure}{.45\textwidth}
          \centering
          \includegraphics[trim={0.25cm 0.35cm 0.3cm 0.26cm},clip,width=.99\linewidth]{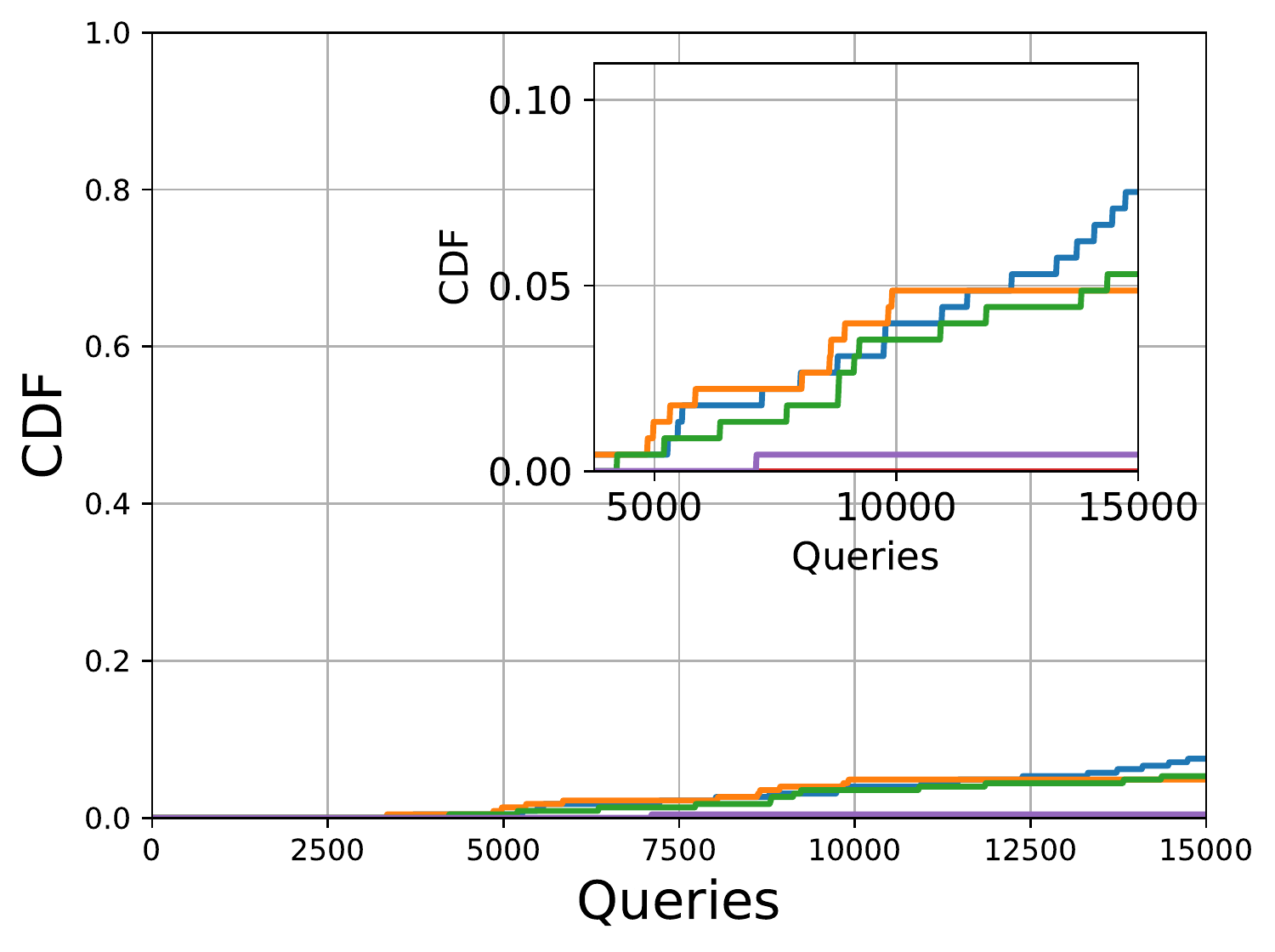}
          \caption{$\varepsilon_\infty=0.01$}
          \label{fig:imagenet_01}
        \end{subfigure}
        \caption{Cumulative fraction of test set images successfully misclassified with adversarial examples generated by GenAttack, Parsimonious, Square, Frank-Wolfe, and our BOBYQA based approaches for different maximum perturbation energies $\varepsilon_\infty$ and DNNs trained on the ImageNet dataset. In all results the solid and dashed lines denoted by `Non-Adv' and `Adv' corresponds to attacks on networks trained without or with the MadryLab defense strategy \cite{robustness} respectively.
        }
        \label{fig:Imagenetcdf}
\end{figure*}

\paragraph{CIFAR10} The CIFAR10 data-set contains images from 10 classes and of dimension 32x32x3. To generate a comprehensive distribution for the queries at each energy budget, ten correctly classified images are consider per each class, and each of them is targeted to all of the 9 remaining classes; this way we generate a total of 900 attacks per maximum perturbation energy per adversarial method.

\paragraph{ImageNet} This data-set contains millions of images with a dimension of 299x299x3 divided among 1000 classes. Because of the high dimensionality and number of classes, random images are attacked considering a random target class. We conducted 200 and 160 tests for networks trained both with and without adversarial training per maximum perturbation energy.

\subsection{Results for Standard and MadryLab Trained DNNs}\label{sec:knwon_experiments}

In Figures \ref{fig:CIFARcdf} and \ref{fig:Imagenetcdf} we present the cumulative fraction of images successfully misclassified (abridged by CDF for cumulative distribution function)  as a function of the number of queries to the DNN for different maximum perturbation energies $\varepsilon_\infty$. The pixels are normalized to be in the interval $(-1/2,1/2)$, hence, $\varepsilon_\infty=0.1$ would imply that any pixel is allowed to change $10\%$ of the total intensity range from its initial value. The CDFs are illustrated so that we can easily see which method has been able to misclassify the largest fraction of images in the given test-set for a fixed number of queries to the DNN.

For the CIFAR10 data-set in Figure \ref{fig:CIFARcdf}, we observe that algorithms that search the perturbation directly in the vertices of the perturbation domain require the least amount of network queries. In the case of non-adversarially trained networks, the Square algorithm is able to misclassify using the least number of queries; this is demonstrated by its associated solid green CDF being consistently above that of the other methods. Specifically, when $\varepsilon_\infty=0.05$, at 1,000 queries Square algorithms has a CDF of 0.97 compared to 0.94 and 0.88 of the Parsimonious and BOBYQA methods respectively, and for $\varepsilon_\infty=0.005$ at 3,000 queries Square achieves a CDF of 0.20 which is $50\%$ times higher than Parsimonious and BOBYQA. When the net is instead trained adversarially, dashed lines, Square algorithm looses a lot of its effectiveness becoming comparable to the BOBYQA based method, while Parismonious algorithm achieves almost always the highest fraction of successfully perturbed images for any given maximum number of queries. For example, when $\varepsilon_\infty=0.05$ at 3,000 queries the CDF of Parisomonious is 0.29 compared to 0.25 and 0.23 of Square and BOBYQA.

In the ImageNet dataset, see Figure \ref{fig:Imagenetcdf}(a), we observe that an adversarial method can be especially susceptible to particular defenses. Specifically, when the network is trained without a defense, the Square algorithm has a success rate CDF that is consistently higher than the other methods, but the success rate CDF for the Square algorithm is decreased by the MadryLab defense so that it is substantially less effective than Parsimonious and BOBYQA algorithms. On the other hand, the Parsimonious method achieves similar results to Square algorithm in the non-adversarial case. On average for the different maximum perturbation energies Parsimonious is 0.045 less efficient than Square, but when the defense is introduced it finds the adversarial examples with the least number of queries. In Figure \ref{fig:Imagenetcdf}(a) Parisomious has a CDF of 0.33 at 15,000 queries while BOBYQA 0.24 and Square 0.07. The rate with which the CDFs decrease as the maximum perturbation energy $\varepsilon_\infty$ decreases it also differs by algorithm. The CDF for Square decreases moderately faster than for Parsimonious such that Square has a consistently higher CDF than Parsimonious for $\varepsilon=0.1$ in Figure \ref{fig:Imagenetcdf}(a) but consistently lower in Figure \ref{fig:Imagenetcdf}(d). Moreover, the success rate for BOBYQA decreases the slowest with $\varepsilon_\infty$ such that in Figure \ref{fig:Imagenetcdf} its CDF is similar to or grater than Parsimonious. Specifically, in Figure \ref{fig:Imagenetcdf}(d) at 15,000 the final CDF of BOBYQA algorithm queries is 1.42 times higher than the one of the Square algorithm.

The Frank-Wolfe algorithm is able to achieve results comparable to the ones of the methods above while considering the small-dimensional problem of CIFAR10 with a very low maximum perturbation energy. However, when considering the ImageNet case and the adversarially trained DNNs, the Frank-Wolfe algorithm has a substantially lower success rate CDF; e.g. in the ImageNet case with non-adversarial training, Square algorithm achieves a CDF 1.66 times higher than the Frank-Wofle algorithm when $\varepsilon_\infty=0.05$.

Finally, GenAttack has a higher success rate CDF than the Frank-Wolfe algorithm in the ImageNet case for $\varepsilon_\infty=0.1$, see Figure \ref{fig:Imagenetcdf}(a), but, besides this case, it constantly achieves the lowest success rate.

\subsection{Results with Fixed Pixel Count Constraints}\label{sec:unkonown_exp}

In addition to network training designed to increase robustness, such as MadryLab considered previously, there are a multitude of other defenses and real world constraints \cite{hao2020adversarial}. The relative success rate, or other characteristics, of adversarial algorithms can be expected to differ in these diverse settings. To demonstrate this, we consider one such setting where the maximum number of pixels allowed to be perturbed is limited. This is motivated by the defenses where network inputs are thresholded in a wavelet domain to exclude high frequency perturbations \cite{guo2018countering}, as well as by real world constraints such as attacks designed to appear structured such as localized perturbations designed to look like graffiti \cite{eykholt2017robust, naseer}.
We allow the algorithms to perturb only the fixed selection of the 1,000 pixels of the targeted image that have the highest variance in intensity in their channel neighborhood. Because of the previous results it is possible to identify three methods that work consistently better than the others, and thus only these will be considered, namely: the Parsimonious, the Square, and the BOBYQA based algorithms. To allow the perturbations to be limited to the selected pixels, we consider the Square algorithm with squares of pixel dimension, the Parsimonious algorithm on the finest grid, and the BOBYQA algorithm without the hierarchical lifting, i.e. $\textbf{D}^1 = \textbf{I}$ where $\textbf{I}$ is the identity matrix. 

\begin{figure*}[t!]
    \centering
        \begin{subfigure}{.45\textwidth}
          \centering
          \includegraphics[trim={0.25cm 0.35cm 0.3cm 0.3cm},clip,width=.99\linewidth]{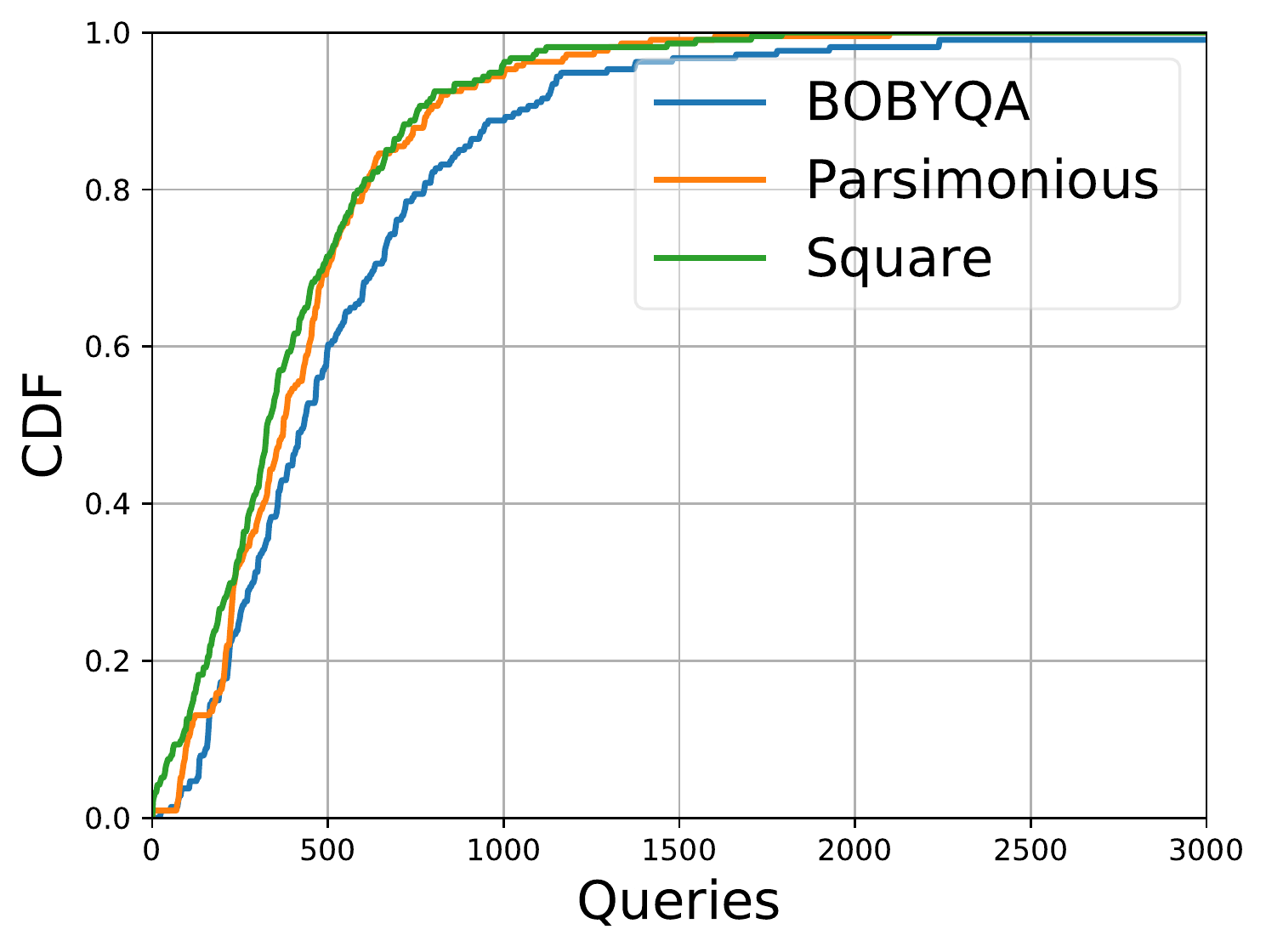}
          \caption{$\varepsilon_\infty=0.2$}
        \end{subfigure}%
        \begin{subfigure}{.45\textwidth}
          \centering
          \includegraphics[trim={0.25cm 0.35cm 0.3cm 0.3cm},clip,width=.99\linewidth]{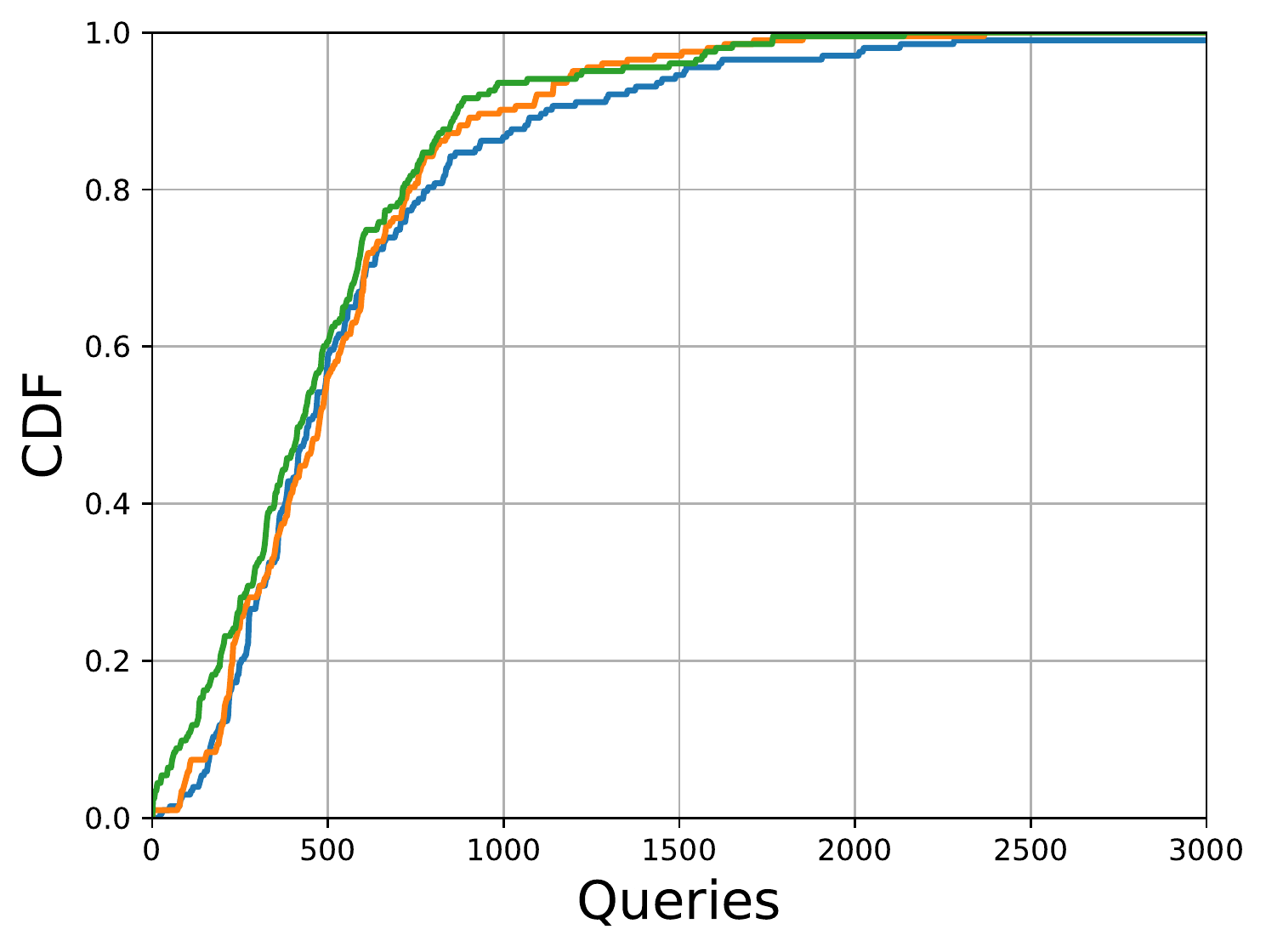}
          \caption{$\varepsilon_\infty=0.15$}
        \end{subfigure} \\
        \begin{subfigure}{.45\textwidth}
          \centering
          \includegraphics[trim={0.25cm 0.35cm 0.3cm 0.3cm},clip,width=.99\linewidth]{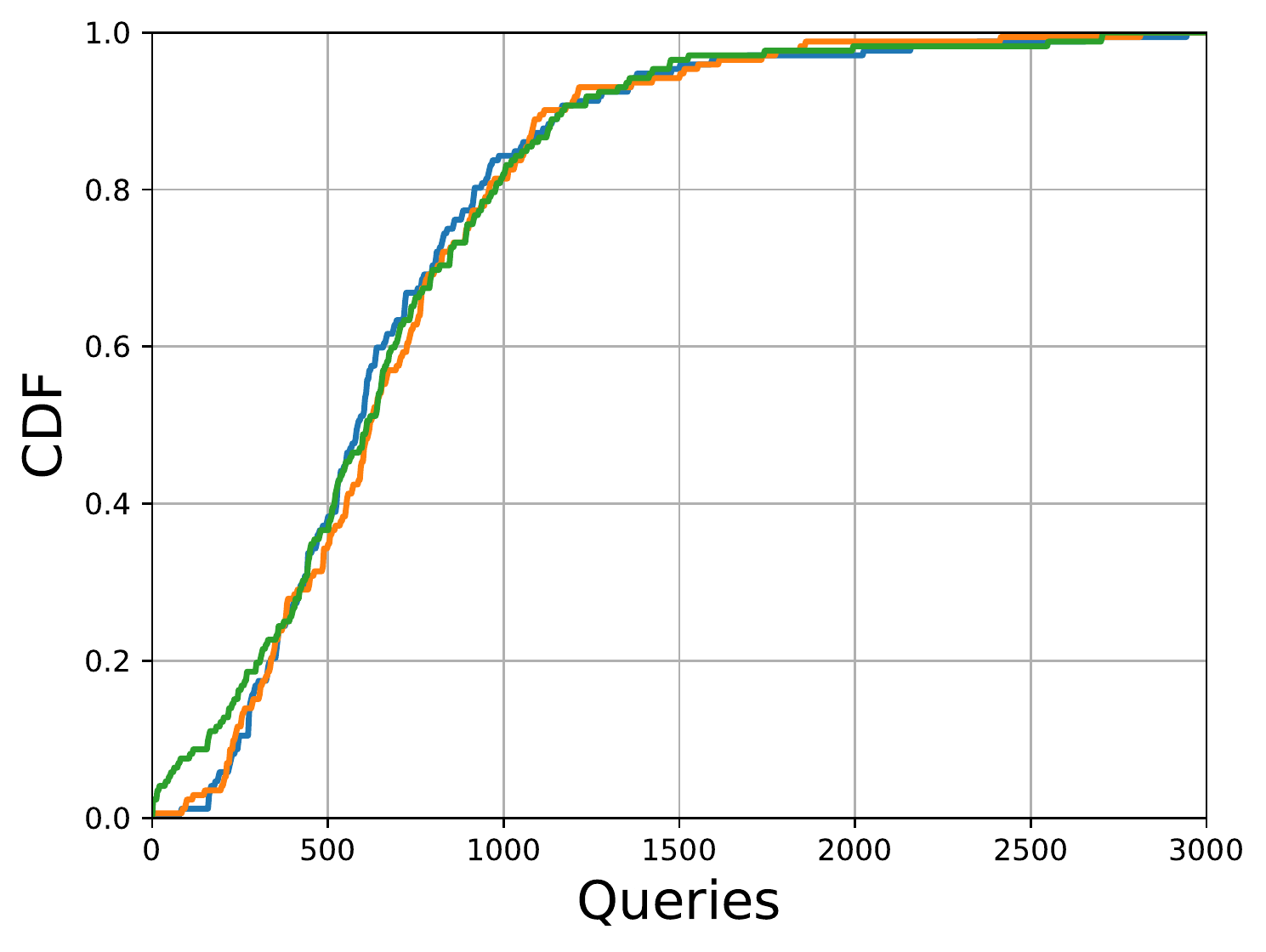}
          \caption{$\varepsilon_\infty=0.1$}
        \end{subfigure}%
        \begin{subfigure}{.45\textwidth}
          \centering
          \includegraphics[trim={0.25cm 0.35cm 0.3cm 0.3cm},clip,width=.99\linewidth]{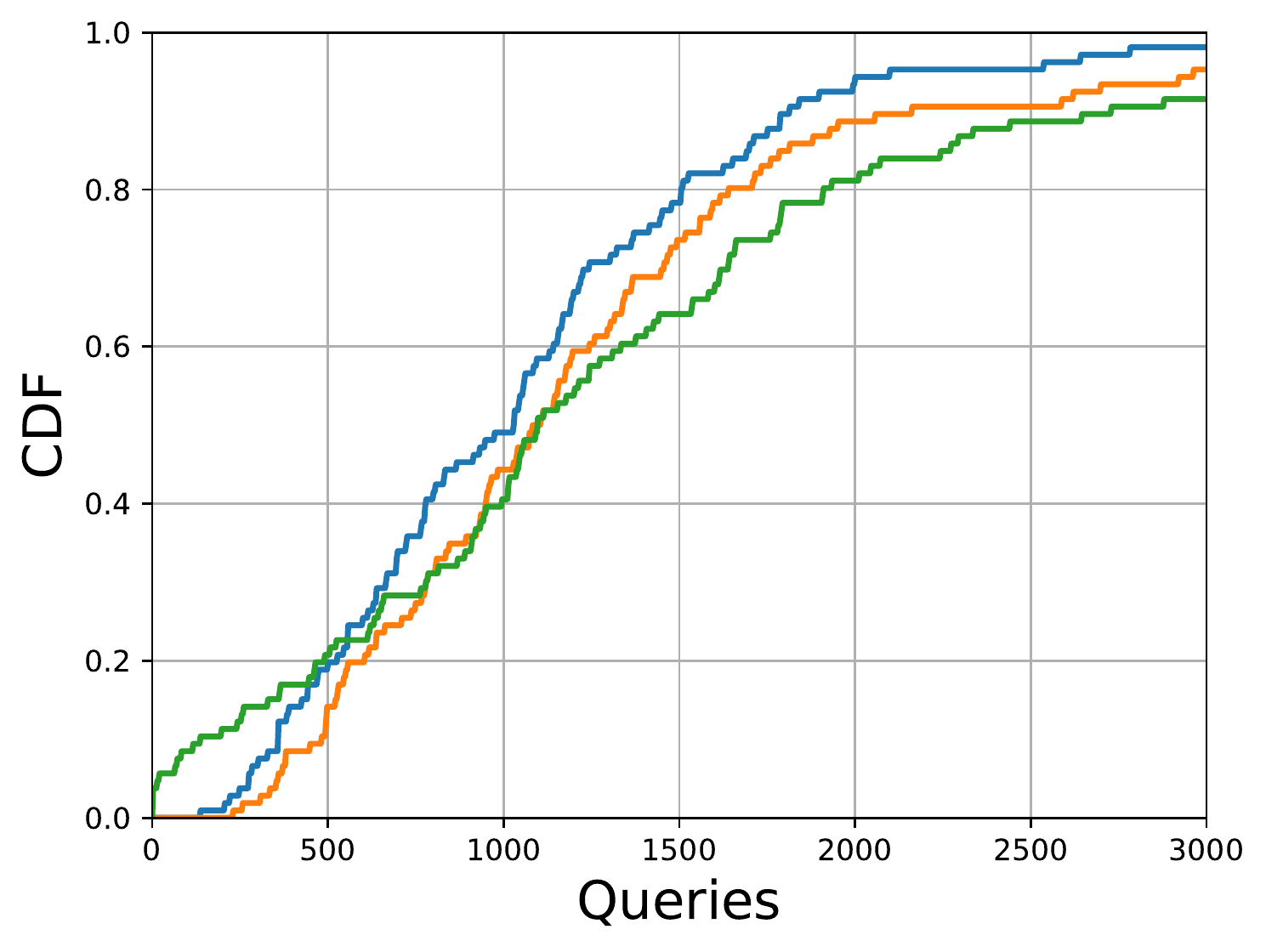}
          \caption{$\varepsilon_\infty=0.05$}
        \end{subfigure}
        \caption{Cumulative fraction of test set images successfully misclassified with adversarial examples generated by Parsimonious, Square, and our BOBYQA based approaches for different maximum perturbation energies $\varepsilon_\infty$ against a ResNet50 trained non-adversarially on the CIFAR10 dataset when only the 1000 pixels with the highest variance in intensity in their neighborhood are allowed to be modified.  
        }
        \label{fig:CIFARcdfreduced}
\end{figure*}

The results reported in Figure \ref{fig:CIFARcdfreduced} suggest that when the domain is dimensionally limited, the most efficient algorithm changes according to the allowed maximum perturbation energy. When the maximum perturbation energy decreases and the linear model is more accurate, the BOBYQA method manages to achieve a higher SR than both Square and Parsimonious algorithms, unlike in the previous experiments. Moreover, the Parsimonious algorithm has almost identical behavior to Square algorithm for high energy bounds, but becomes more efficient when the maximum energy is $\varepsilon_\infty=0.05$. We also considered experiments on ImageNet, but limiting the number of pixels that could be perturbed did not allow for any successful misclassification with less than 15,000 queries.

\section{Discussion and Conclusion}
\label{sec:summary}

We have compared for the first time how the the existing GenAttack \cite{Alzantot}, Parsimonious \cite{COMBI}, Square \cite{andriushchenko2019square}, and Frank-Wolfe \cite{chen2020frank} algorithms, and the newly introduced BOBYQA based method, behave when the available $\ell^\infty$ energy for a perturbation varies, and an adversarial training or a structural defense is considered.

The results suggest that those methods limiting the search for an adversarial example to the vertices of the $\ell^\infty$ perturbation domain generally work better. Whilst Square algorithm is especially effective on the non-adversarially trained networks, the Parsimonious algorithm manages to outperform any other approach when the networks are adversarially trained with the MadryLab implementation. Furthermore, the Parsimonious algorithm performs better than Square when considering the structural defense that limits the attacks on some pixels, suggesting that an algorithm based on combinatorial search is robust in its hyper-parameters to the setting where it is applied. 

The BOBYQA based algorithm was introduced in this paper to explore how model-based approaches compare to the state-of-the-art algorithms, and was found to achieve similar results to the Parsimonious and Square algorithms. In almost in all the experiments the BOBYQA based algorithm achieves a success rate CDF comparable to the ones of the Parsimonious and the Square algorithms; it achieves the state-of-the-art success rate at saturation for low maximum perturbation energy constraint both in the ImageNet case and in the pixel constrained problem. Moreover, new dimensionality reduction techniques that are being considered in DFO, see for example \cite{scalable_DFO}, might improve the results observed here and lead to a state-of-the-art algorithm for the generation of adversarial examples.

In conclusion, we find that both the structure of the algorithm and the attack setting have the potential to impact the algorithm performance. These observations highlight the importance of comparing any new algorithm to the state-of-the-art in a variety of different settings, such as is done here. Similarly, the effectiveness of an adversarial defense for DNNs should always be tested using as wide a range of algorithms as possible.

\bibliographystyle{spmpsci}      
\bibliography{bibliography}   

%
%

\clearpage

\end{document}